\documentclass[pdflatex,sn-mathphys-num]{sn-jnl}

\usepackage{graphicx}%
\usepackage{multirow}%
\usepackage{amsmath,amssymb,amsfonts}%
\usepackage{amsthm}%
\usepackage{mathrsfs}%
\usepackage[title]{appendix}%
\usepackage{xcolor}%
\usepackage{textcomp}%
\usepackage{manyfoot}%
\usepackage{booktabs}%
\usepackage{algorithm}%
\usepackage{algorithmicx}%
\usepackage{algpseudocode}%
\usepackage{listings}%
\usepackage{comment}

\usepackage{array}
\usepackage{subcaption}
\usepackage{arydshln}

\usepackage[whole]{bxcjkjatype} 

\usepackage{booktabs}
\usepackage{multirow}
\usepackage{makecell}

\usepackage{siunitx}

\theoremstyle{thmstyleone}%
\theoremstyle{thmstyletwo}%

\theoremstyle{thmstylethree}%

\begin{document}

\title[Cybernetic Android Avatar ``Yui'': System Integration, Field Deployment, and Evaluation]{Cybernetic Android Avatar ``Yui'': System Integration, Field Deployment, and Evaluation}

\author[1]{\fnm{Kaoruko} \sur{Shinkawa}}\email{kshinkawa@uec.ac.jp}
\equalcont{These authors contributed equally to this work.}

\author[2]{\fnm{Mizuki} \sur{Nakajima}}\email{mizuki.nakajima@mail.dendai.ac.jp}
\equalcont{These authors contributed equally to this work.}

\author[1]{\fnm{Taisei} \sur{Mogi}}\email{taisei.m24@gmail.com}

\author*[1]{\fnm{Yoshihiro} \sur{Nakata}}\email{ynakata@uec.ac.jp}
\equalcont{These authors contributed equally to this work.}

\affil*[1]{\orgdiv{Department of Mechanical and Intelligent Systems Engineering, Graduate School of Informatics and Engineering}, \orgname{The University of Electro-Communications}, \orgaddress{\street{1-5-1 Chofugaoka}, \city{Chofu}, \postcode{182-8585}, \state{Tokyo}, \country{Japan}}}

\affil[2]{\orgdiv{Department of Mechanical Engineering, Graduate School of Engineering}, \orgname{Tokyo Denki University}, \orgaddress{\street{5 Senjuasahicho}, \city{Adachi-ku}, \postcode{120-8551}, \state{Tokyo}, \country{Japan}}}

\abstract{Remote communication technologies have become widely used; however, supporting a sense of shared physical space and conveying rich non-verbal cues remain challenging in many social interaction scenarios. This study presents ``Yui,'' a full-body cybernetic android avatar designed to integrate operator-side immersive teleoperation with interlocutor-side human-like social signaling. Yui combines a 55-degrees of freedom full-body mechanism with a previously developed android head, facial expression and gaze control, upper-body and arm motion, hand actuation, and a mobile platform. It can be operated through either the immersive mode using a head mounted display-based interface or desktop mode using a webcam-based interface. We evaluated the system through three real-world deployments: a long-term public exhibition at Expo 2025 in Osaka, Kansai, Japan; a remote educational exchange between elementary school students; and a public interaction study with general participants. During the Expo deployment, two units accumulated approximately 1131 h of operation, demonstrating both operational feasibility and maintenance challenges. In the public study, both operators and interlocutors reported positive impressions of co-presence and willingness to use the system. Interlocutors also rated the avatar positively in terms of human likeness and the transmission of emotions and intentions. The results indicate usability for general operators while suggesting room for improvement in precise controllability. These findings provide field-derived evidence and design implications for socially deployable full-body android avatars.}


\maketitle

\section{Introduction}\label{sec1}
Following the COVID-19 pandemic, video conferencing has become a critical communication infrastructure. 
However, the current systems still fall short of face-to-face interactions. 
Bailenson \cite{Bailenson2021} reported that excessive eye contact and restricted movement in video conferencing impose substantial cognitive load, a phenomenon known as ``Zoom Fatigue.'' 
To mitigate such constraints, telepresence robots that provide remote operators with spatial mobility have been increasingly adopted, with reported benefits in domains that require trust and intimacy, including elderly care \cite{Beer2011, Cesta2016}, education \cite{Newhart2017, Tanaka2014, Lei2022}, and family communication \cite{Seo2024}. 
Nevertheless, the media richness theory (MRT) \cite{Daft1986} and notion of common ground \cite{Clark1991} suggest that high-quality social interaction relies not only on information exchange but also on rich non-verbal cues and a shared physical space (co-presence) \cite{Li2015}. 
Although telepresence systems enable mobility, most interfaces remain limited to flat, two-dimensional displays, which can weaken the perceived embodiment in three-dimensional space and reduce the effectiveness of non-verbal cues, such as gaze and body orientation. 
Consequently, reproducing a social presence comparable to that in face-to-face encounters remains challenging in interaction scenarios that demand nuanced interpersonal communication \cite{Deng2019}.

To address this ``embodiment gap,'' research on physically embodied avatar robots has intensified \cite{Darvish2023, Zhang2022}. 
Existing approaches can be broadly categorized into two types. 
The first prioritizes the interlocutor's experience by pursuing human-likeness in appearance and behavior. 
The second prioritizes the operator's experience by emphasizing immersive perception in a remote environment and physical interactions, including manual tasks. 
Achieving both at a high level within a single system remains difficult. 
Interlocutor-centric androids often emphasize appearance while providing limited support for the operator's sense of embodiment, whereas operator-centric avatars may offer limited support for an emotional connection with the interlocutor. 
This study aims to design and socially implement a full-body android avatar that integrates these two directions.

The first direction uses human-like robots (androids) to examine how appearance and motion fidelity shape an interlocutor's experience \cite{Sato2022}. 
Recent progress is evident in commercial platforms, such as Ameca \cite{Ameca2025} and Sophia \cite{Sophia2025}, as well as in academic systems, including the autonomous conversational android Erica \cite{Glas2016}, android EveR-4 \cite{Ahn2012}, and social robot Nadine \cite{Thalmann2017}. 
Earlier studies have extended Android technologies to teleoperated avatars, including Geminoid \cite{Nishio2007} and Telenoid \cite{Ogawa2011}. 
Sakamoto et al. \cite{Sakamoto2007} showed that teleoperated androids can convey social presence to the interlocutor by approximating the operator's physical presence. 
However, many teleoperated-android studies have focused primarily on the interlocutor, whereas the operator interface often remains desktop-based. 
Even when speech-driven lip synchronization is available, the operators do not necessarily share the avatar's field of view from an immersive perspective. 
Consequently, body ownership grounded in a first-person viewpoint has not been fully addressed, and enabling bidirectional immersive interaction, in which the operator perceives the avatar as their own body while sustaining natural eye contact, remains an open challenge.

The second direction focuses on maximizing the operator's immersion and task performance in remote environments. 
Representative systems include TELESAR VI \cite{Tachi2020}, which uses telexistence to transmit physical interaction, including haptic feedback; iCub3 \cite{Dafarra2024}, which enables whole-body teleoperation with multimodal feedback; and platforms such as NimbRo \cite{Lenz2025}, AVATRINA \cite{Correia2025}, and Alter-Ego X \cite{Zambella2025}, which demonstrated strong performance in the ANA Avatar XPRIZE competition. 
By integrating virtual reality (VR) and haptic technologies, these systems enable operators to navigate remote spaces and perform dexterous tasks. 
However, they generally prioritize functionality and task efficiency, with social expressiveness remaining secondary.
For example, NimbRo \cite{Lenz2025} uses a physically flat screen to represent the face, whereas iCub3 \cite{Dafarra2024} employs LED-based symbolic facial expressions. 
Although effective for task execution, such designs may be limited to conveying gaze-related feedback and empathetic facial expressions, which are important in contexts such as elderly care and intimate dialogues. 
Moreover, although JANUS \cite{CisnerosLimon2025} uses the humanoid platform HRP-4C \cite{Kaneko2009}, its system integration emphasizes high-level skill transfer and connectivity rather than the fidelity of social interaction. 
Thus, despite strong task performance, limitations remain in reproducing social presence and communicating nuanced affect, which this study aims to address.

Against this background, we propose a full-body android avatar, ``Yui,'' that integrates these two directions by transmitting human-like social signals to the interlocutor while providing the operator with an immersive first-person experience and a sense of body ownership.
Our primary objective is to improve the quality of social interaction in remote environments rather than to focus solely on task-based functionality. 
This approach is consistent with the cybernetic avatar (CA) concept \cite{Ishiguro2025}, which aims to facilitate remote activities within the context of interpersonal communication. 
Building on the high-fidelity head unit developed in our previous study \cite{Nakajima2024}, this study describes its extension to a full-body interaction system. 
In particular, we integrate an arm and mobility mechanism \cite{Nakata2022} to support natural human-like movements, together with a teleoperation interface that builds on our previous vision-sharing approach \cite{Shinkawa2024} and reflects the operator's gaze and body motions.
As body movements and postures are as important as facial expressions for conveying affective states \cite{Kleinsmith2013}, full-body degrees of freedom (DOF) are necessary to reproduce social presence. 
In addition, gaze is a key non-verbal cue for communicating engagement in human--robot interactions (HRI) \cite{Admoni2017, Vazquez2017}. 
Accordingly, we consider the natural reproduction of the operator's field of view and gaze behavior to be an important factor in sustaining natural interaction and trust in the interlocutor.
By integrating these components, the proposed system enables natural eye contact and gesturing through intuitive teleoperation without specialized training, thereby supporting the reconstruction of an operator's social presence.

This paper presents the system configuration of the full-body android avatar ``Yui,'' integrating the previously proposed head unit with a whole-body mechanism for human-like motion and a mobility platform. 
To support whole-body control, we describe an immersive teleoperation system based on head-mounted display (HMD) tracking. 
We further present a vision-based remote head-operation interface using a standard webcam for users who may not be able to wear an HMD, such as children. 
Our evaluation emphasizes utility and acceptance in real-world settings (``in the wild'') rather than in controlled laboratory studies.
Accordingly, this study adopts an exploratory field-oriented evaluation approach rather than a hypothesis-driven experimental design.
As \v{S}abanovi\'{c} et al. \cite{Sabanovic2006} argued, a real-world evaluation with a high ecological validity is necessary to assess the social acceptance of social robots. 
We evaluated technical robustness and social acceptance through deployments across multiple field settings, including public events, educational environments, and long-term operations, at Expo 2025 in Osaka, Kansai, Japan. 
These evaluations were based on questionnaire surveys and operational data collected during the deployments. 
Based on these field-derived insights, we discuss practical design guidelines and challenges for future social implementation of full-body android avatars.

The main contributions of this study are as follows:
\begin{enumerate}
    \item \textbf{Development of a full-body android avatar system:} Integration of a previously developed head unit with a whole-body mechanism comprising the torso, arms, and mobility units, capable of reproducing human-specific movements such as those of the scapulae and knees.

\item \textbf{Integration of an immersive teleoperation interface:} Development of a system combining HMD-based tracking (gaze, head, and hands) with vision-based facial expression recognition, designed to facilitate intuitive whole-body control and a sense of body ownership.

\item \textbf{Long-term real-world deployment:} Evaluation of technical robustness and operational stability through a six-month deployment at events associated with Expo 2025 Osaka, Kansai, Japan (Study 1).

\item \textbf{Assessment of social acceptance across diverse scenarios:} Verification of social utility and acceptance through interaction studies with the general public in real-world settings, including educational environments (Study~2) and public events (Study~3).

\item \textbf{Design guidelines for social implementation:} Provision of practical guidelines for the hardware design and operation of full-body android avatars, derived from empirical insights gained through field deployments.
\end{enumerate}

\section{Android Avatar: Concept and System Overview}\label{sec2}
This section outlines the concept of the proposed android avatar ``Yui'' and presents an overview of the system. This study addresses the limitations identified in Section~\ref{sec1}, specifically, the lack of embodiment and non-verbal cues in remote communication. Our approach employs a full-body android avatar as the physical medium to support seamless social interaction between a remote operator and local interlocutor.

\subsection{Concept: Toward Seamless Operator--Interlocutor Engagement}\label{sec21}

\begin{figure}[t]
  \centering
  \includegraphics[width=1.0\textwidth]{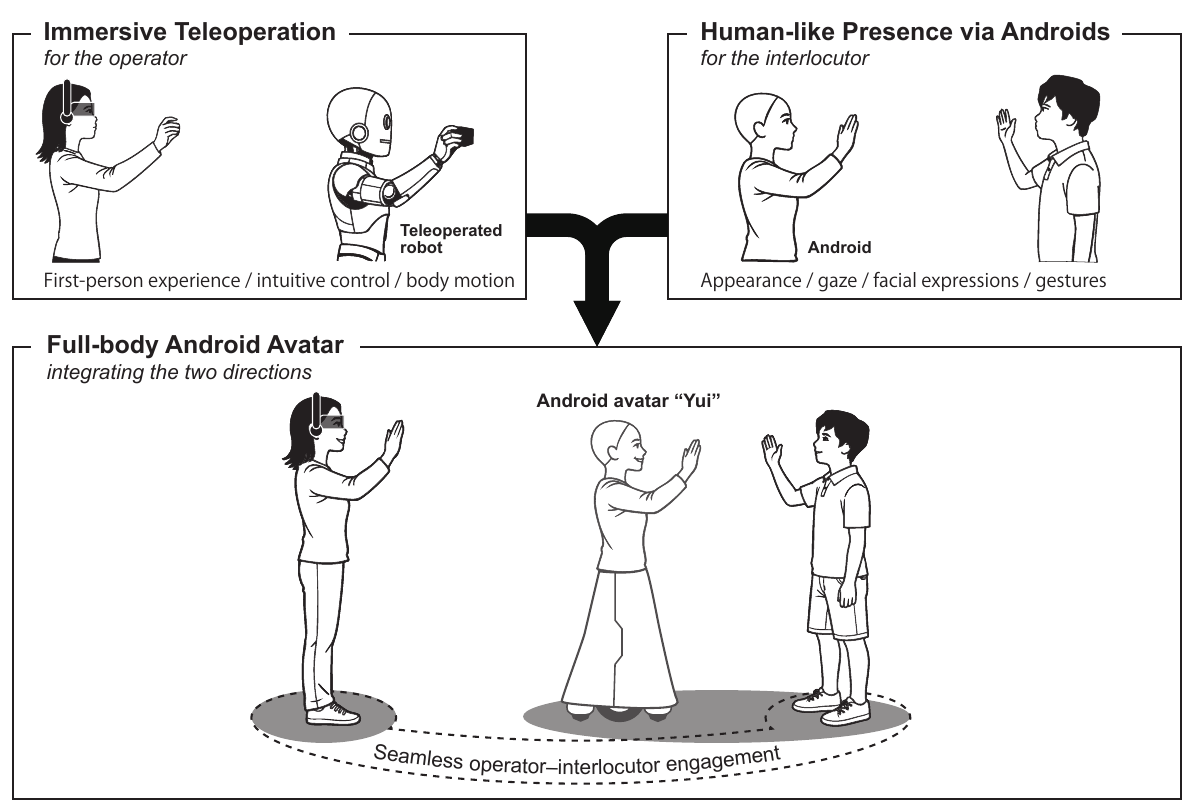}
  \caption{Conceptual positioning of the full-body android avatar approach. The concept combines operator-centered immersive teleoperation with interlocutor-centered human-like presence and social signaling to enable seamless engagement between an operator and interlocutor.}
  \label{fig:yui_concept}
\end{figure}

The concept of an android avatar is shown in Fig. ~\ref{fig:yui_concept}. The related studies can be broadly divided into two categories. The first is the avatar and teleoperation research, which enables operators to access remote environments with an emphasis on immersion and operability. The second is android research, which investigates the interlocutor's experience through a human-like appearance and expressive behaviors. This study aims to integrate these two directions using the full-body android avatar, Yui. This integration is intended to combine the operator's first-person experience with social cues presented to the interlocutor.

In this paper, ``seamless engagement'' refers to an interactional state in which the avatar does not become a prominent mediator and the exchange between the operator and interlocutor proceeds naturally. To support this, the operator receives audiovisual input from the avatar from a first-person perspective, while the operator's facial expressions, gaze, head pose, upper-body motion, and arm and hand movements are mapped onto the avatar. On the interlocutor side, multiple social signals such as gaze, facial expressions, gestures, posture, and spoken dialogue are presented. The goal is to facilitate interaction that can be experienced as if it were face-to-face, despite physical separation.

\subsection{System Overview}\label{sec22}

\begin{figure}[t]
  \centering
  \includegraphics[width=1.0\textwidth]{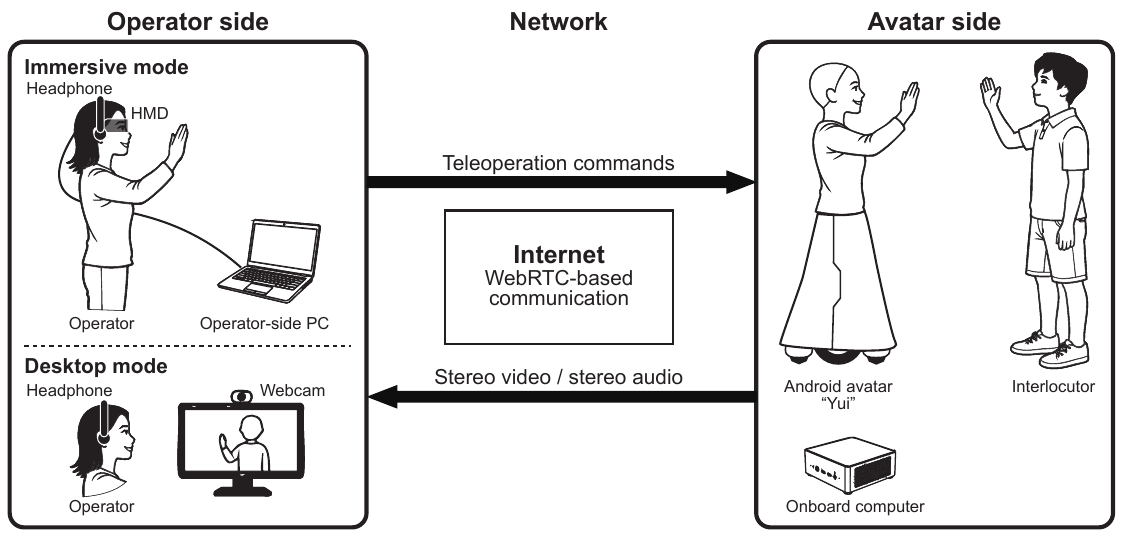}
  \caption{Overview of the teleoperation system. The operator side supports both immersive and desktop modes. Teleoperation commands are transmitted from the operator side to the avatar side via web real-time communication (WebRTC)-based communication, while stereo video and audio are transmitted back to the operator.}
  \label{fig:sys_overview}
\end{figure}

An overview of the proposed system is shown in Fig. ~\ref{fig:sys_overview}. The system consists of three components: the avatar side, operator side, and network connecting them.

\begin{description}
    \item[\textbf{Avatar side}] The avatar side comprises the full-body android avatar (i.e., Yui) and an avatar-side PC. The avatar-side PC controls the entire body based on information transmitted from the operator and simultaneously streams the sensor data from the avatar, including stereo video and audio, to the operator.
    \item[\textbf{Operator side}] The operator side comprises an HMD and an operation PC. It captures the operator's facial expressions, gaze, head pose, upper-body motion, and arm and hand movements. It also receives and presents the stereo camera video and stereo microphone audio, which are provided as spatial audio streamed from the avatar's side.
    \item[\textbf{Network}] For bidirectional streaming and teleoperation, the system uses WebRTC. From the operator side to the avatar side, tracking information, including facial expressions, gaze, head pose, upper-body motion, arm and hand motion, and mobility commands, are transmitted. Stereo video and audio are continuously streamed from the avatar's side to the operator side.
\end{description}

As shown in Fig.~\ref{fig:sys_overview}, the system supports two operator interfaces: Immersive and desktop modes. In addition, an auxiliary control interface and game controller can be connected for system startup, debugging, and simplified operation. The desktop mode is based on MediaPipe~\cite{CLugaresi2019} and enables users who cannot wear an HMD to perform remote head operations. This mode recognizes the facial expressions and head movements during teleoperation. In the interaction evaluations reported in this study, we primarily used the immersive mode.

\section{Android Avatar Hardware}\label{sec3}
This section describes the hardware design of the android avatar, Yui. 
We first present the mechanical design, including the degrees-of-freedom (DoF) layout and configuration of each subsystem. 
We then describe the electrical design, including the power supply, control electronics, sensing systems, and audio devices.

\subsection{Mechanical Design}\label{sec31}
\begin{figure}[t]
  \centering
  \includegraphics[width=1.0\textwidth]{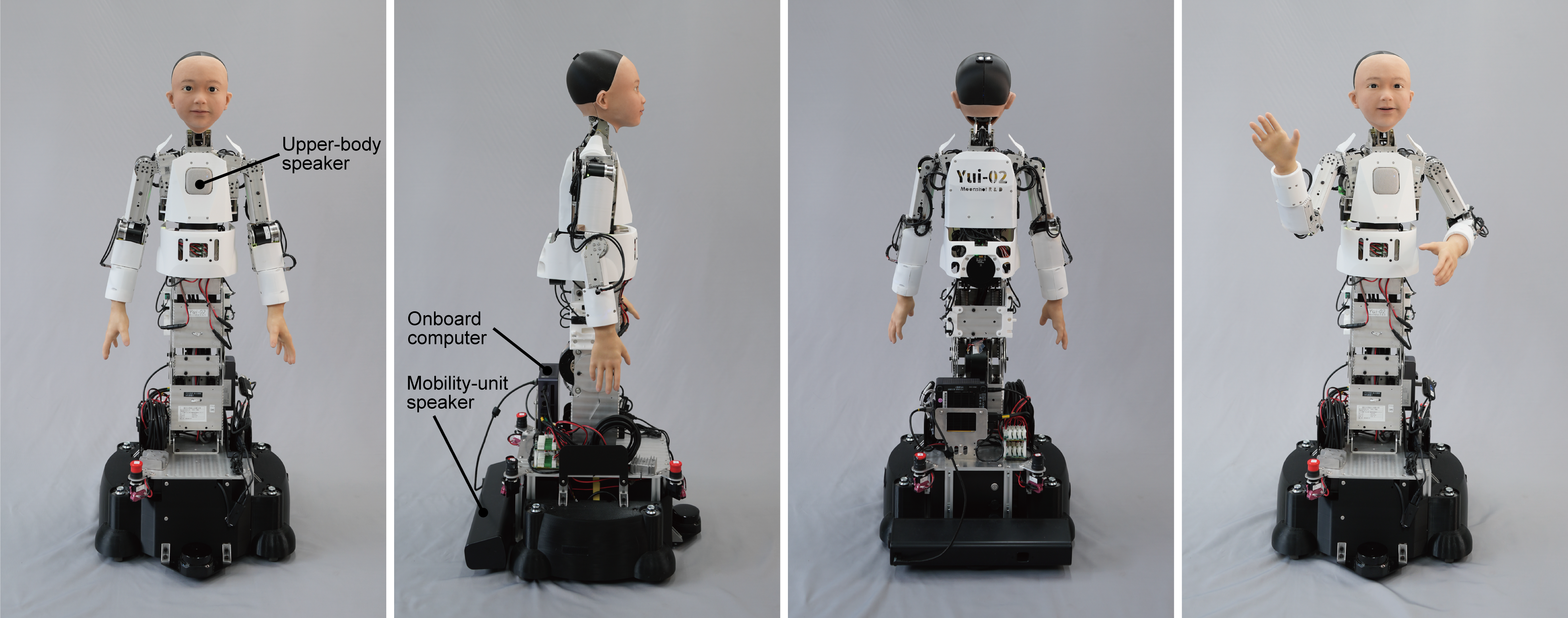}
  \caption{Appearance of the full-body android avatar ``Yui,'' showing front, side, and rear views, along with an example posture demonstrating arm and hand motion. The labeled views indicate the locations of the onboard computer and two speakers used for audio output.}
  \label{fig:yui_appearance}
\end{figure}

\begin{figure}[t]
  \centering
  \includegraphics[width=0.8\textwidth]{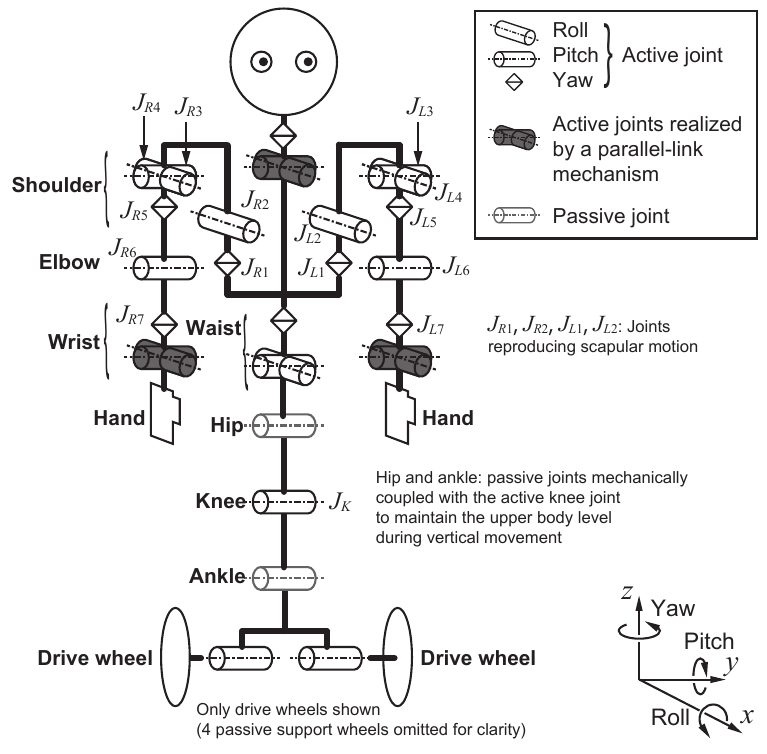}
  \caption{Kinematic structure of Yui. Active joints implemented with parallel-link mechanisms and passive joints are indicated.}
  \label{fig:yui_structure}
\end{figure}

An overview of Yui is presented in Fig. ~\ref{fig:yui_appearance}. The key specifications, including the dimensions, weight, and total DoF, are summarized in Table~\ref{tab:yui}. 
The kinematic structure is illustrated in Fig. ~\ref{fig:yui_structure}. The DoF breakdown by body part is summarized in Table~\ref{tab:yui_dof}. 
The following subsections describe the head unit, body and arms, hands and wrists, and the mobile platforms.

\begin{table}[t]
    \centering
    \renewcommand{\arraystretch}{1.2}
    \caption{Key specifications of ``Yui''}
    \label{tab:yui}
    \begin{tabular}{@{}l l l@{}}
        \toprule
        \multicolumn{2}{@{}l}{Height} & 1200 {[}mm{]} \\
        \multicolumn{2}{@{}l}{Weight including battery} & 43.6 {[}kg{]} \\
        \multicolumn{2}{@{}l}{Onboard computer} &
        \makecell[l]{UM790 Pro (MINISFORUM)\\
        64 GB RAM, Windows} \\
        \multicolumn{2}{@{}l}{Network} &
        \makecell[l]{Wireless network connection via Wi-Fi} \\
        \midrule

        \multirow{6}{*}{Actuators}
        & Head and neck &
        \makecell[l]{DC motors (Maxon Motor AG)} \\
        & Arms &
        \makecell[l]{Brushless DC (BLDC) motors with cycloidal reduction gears\\
        (Keigan Inc.; Sumitomo Heavy Industries, Ltd.)} \\
        & Wrists &
        \makecell[l]{Linear actuators\\
        (P8-25-50-12-P, Actuonix Motion Devices Inc.)} \\
        & Hands &
        \makecell[l]{DC motors (Maxon Motor AG)} \\
        & Waist and knee &
        \makecell[l]{BLDC servo motors with planetary gearboxes\\
        (MG8010-36 Duo, Shanghai Lingkong Technology Co., Ltd.)} \\
        & Mobility unit &
        \makecell[l]{Modified mobile base (Vstone Co., Ltd.)\\
        BLDC in-wheel motors, 40 W $\times$ 2} \\

        \midrule
        \multirow{2}{*}{Motor control}
        & CAN &
        \makecell[l]{Custom controller area network (CAN) motor driver boards\\
        for the head, neck, wrists, and hands} \\
        & RS485 &
        \makecell[l]{Motor control for the arms, waist, knee,\\
        and mobility unit} \\

        \midrule
        \multirow{2}{*}{Sensors}
        & Joint &
        \makecell[l]{Sensors for measuring joint angles\\
        and linear actuator positions} \\
        & Head &
        \makecell[l]{- USB cameras $\times$ 2\\
        \quad (OV5640 sensors with 200$^\circ$ wide-angle lenses)\\
        - Stereo microphones\\
        \quad (CS-10EM, Roland Corporation)} \\

        \midrule
        \multirow{2}{*}{Audio}
        & Upper body &
        \makecell[l]{USB speakerphone\\
        (AT-CSP1, Audio-Technica Corporation)} \\
        & Mobility unit &
        \makecell[l]{Speaker\\
        (Sound Blaster GS3, Creative Technology Ltd.)} \\

        \midrule
        \multirow{2}{*}{Power}
        & Battery &
        \makecell[l]{LiFePO$_4$ battery, 24 V, 60 Ah $\times$ 1} \\
        & Distribution &
        \makecell[l]{Single-battery power system\\
        with DC-- converters for onboard devices} \\

        \midrule
        \multicolumn{2}{@{}l}{Auxiliary input} &
        \makecell[l]{Game controller for startup, debugging,\\
        and simplified operation} \\

        \botrule
    \end{tabular}
\end{table}

\begin{table}[t]
    \centering
    \renewcommand{\arraystretch}{1.2}
    \caption{Degrees of freedom of Yui}
    \label{tab:yui_dof}
    \begin{tabular}{@{}l l@{}}
        \toprule
        Component & DOFs \\
        \midrule
        Total & 55 \\
        \midrule
        Head & 18 \\
        \hspace{1em}Eyes & 3 \\
        \hspace{1em}Face & 15 (22 actuation points)$^{\dagger}$ \\
        Neck & 3 \\
        Arms & $7 \times 2$ \\
        Wrists & $2 \times 2$ \\
        Hands & $5 \times 2$ \\
        Waist & 3 \\
        Mobility unit & \makecell[l]{Knee: 1 \\ Wheels: $1 \times 2$} \\
        \bottomrule
    \end{tabular}

    \vspace{1mm}
    \footnotesize
    $^{\dagger}$ The face has 15 DOFs and 22 facial skin actuation points; 
    the value in parentheses indicates the number of actuation points. 
    For some DOFs, the forward and reverse rotations of a rotary motor actuate different facial skin points.
\end{table}

\subsubsection{Head unit}\label{sec311}
\begin{figure}[t]
  \centering
  \includegraphics[width=\textwidth]{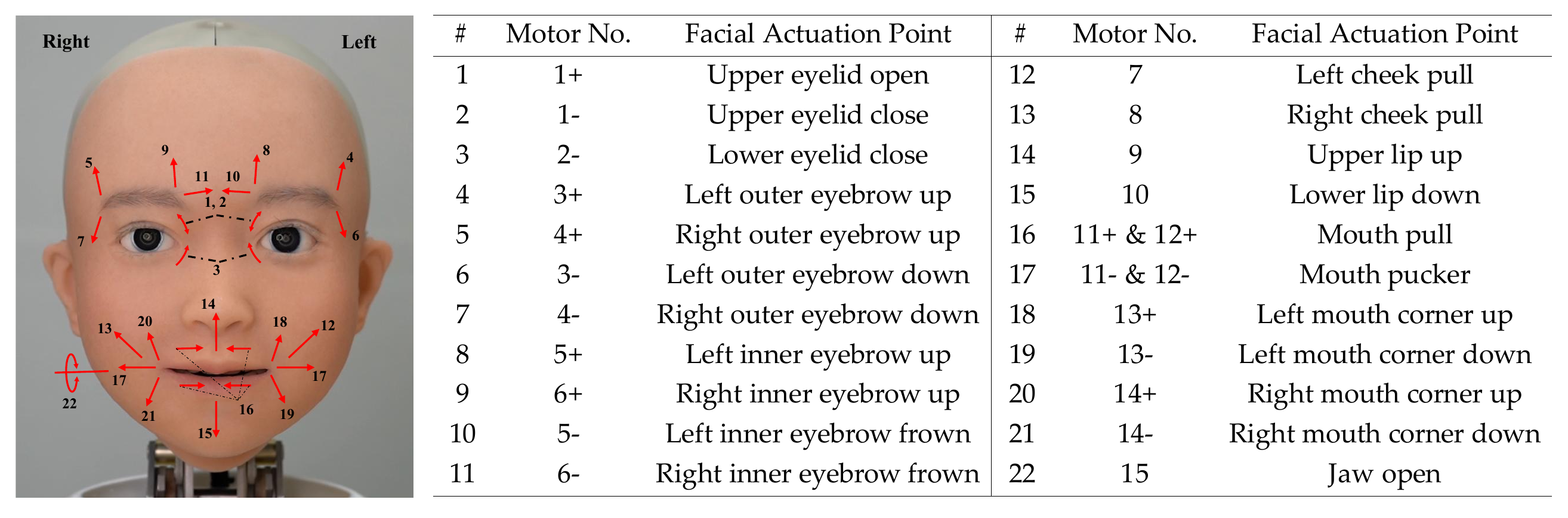}
  \caption{Facial actuation points and motors of Yui, where + and − denote forward and reverse motor rotation, respectively.}
  \label{fig:facial_act_pts}
\end{figure}

For the head, we adopted the appearance and facial-expression-generation mechanism of the android head developed in our previous study \cite{Nakajima2024} as the base design, with modifications to some degrees of freedom of the facial expression.  
The correspondence between the DoF of the facial expression and the motors in the android head used in this study is shown in Fig. ~\ref{fig:facial_act_pts}.
We defined an actuation point as a functional unit representing a type of local displacement that can be generated relative to a neutral facial expression.
This is not a geometric point but an operational unit that represents a mode of change from the neutral state within a specific facial region. 
For the same type of deformation within the same region, a single actuation point is defined, regardless of the direction of displacement. 
In this mechanism, 22 actuation points are realized using 15 motors.

The android head has three degrees of freedom, which enable rotation about the roll, pitch, and yaw axes.
The eyes have three degrees of freedom: two for independent horizontal rotation of the left and right eyes, and one for synchronized vertical rotation.
These drive mechanisms are the same as those described in our previous study~\cite{Nakajima2024}.
The range of motion of the eyes was defined with the forward-facing direction of the lenses as $0^\circ$, with a horizontal range of $\pm 35^\circ$ and vertical range of $-14^\circ$ to $8^\circ$.

\subsubsection{Body and arms}\label{sec312}
The body has three DoF at the waist that enable upper-body posture changes. The waist rotation axes are orthogonal and arranged from bottom to top as the pitch, roll, and yaw, as shown in Fig. ~\ref{fig:yui_structure}. All three waist DoF are driven by the same motor unit (MG8010-36 Duo, Shanghai Lingkong Technology Co., Ltd.), which integrates a brushless DC (BLDC) motor with a planetary gearbox (36:1).

Each arm has seven DoF, consisting of two DoF for reproducing the scapular motion.
Three DoF are at the shoulder, one DoF at the elbow, and one DoF for forearm pronation--supination.
As shown in Fig.~\ref{fig:yui_structure}, the joints labeled $J_{R1}$, $J_{R2}$, $J_{L1}$, and $J_{L2}$ correspond to the scapular motion joints.
The two DoF at the wrist are driven by hand motor drivers, as described in Section~\ref{sec313}.

The arm motor units were developed for the android avatar and consisted of BLDC motors coupled with cycloidal drives. We adopted a cycloidal drive because it uses a contact mechanism different from gear meshing, which can help reduce mechanical noise. Mechanical noise is undesirable not only because it may introduce incongruity in a human-looking android but also because it can interfere with spoken audio presented to the interlocutor and with microphone input during dialogue.

\subsubsection{Hands and wrists}\label{sec313}
\begin{figure}[t]
  \centering
  \includegraphics[width=1.0\textwidth]{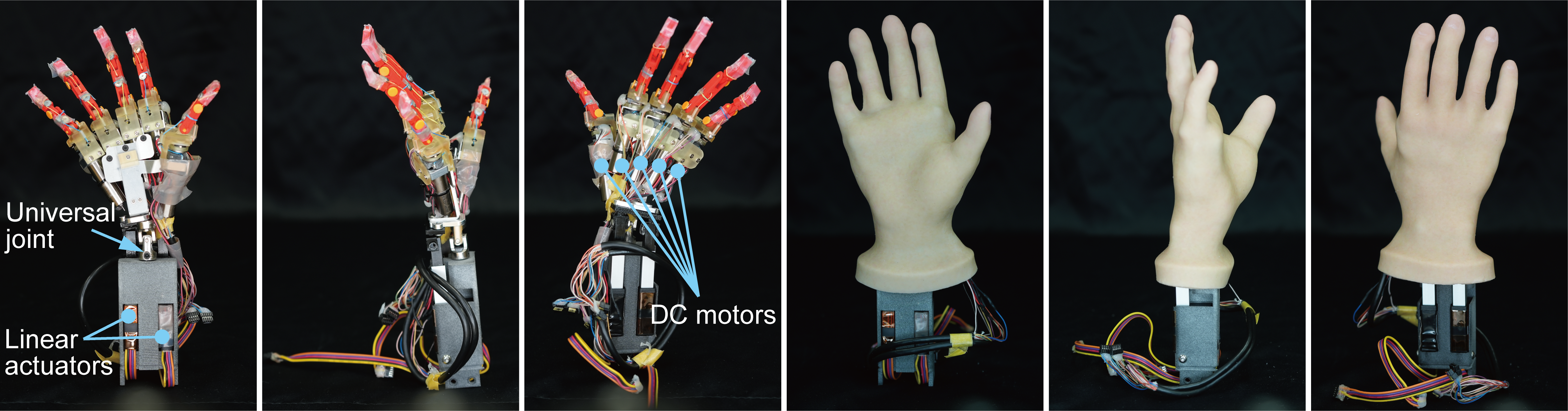}
  \caption{Hand and wrist mechanisms of Yui. The three images on the left show the internal mechanism, and the three images on the right show the appearance with the silicone skin. Within each set, front, side, and rear views are presented from left to right.}
  \label{fig:yui_hand}
\end{figure}

Fig.~\ref{fig:yui_hand} shows the hand and wrist mechanisms, including the internal mechanism and appearance of the silicone skin. Both the hands and wrists are covered with the same silicone skin as the head unit. As a reference for the scale, the length from the tip of the index finger to the wrist is \qty{140}{\milli\meter}. Each hand has five DoF, with one motor unit (DCX10S+GPX10A, maxon) assigned to each finger, including the thumb, to generate flexion via a tendon-driven mechanism. Flexion is produced by pulling the palmar-side tendons, whereas extension is achieved passively using mechanical springs placed on the dorsal side. The motor unit consists of a DC motor and planetary gearbox with a 16:1 gear ratio.

The wrists employ a parallel-link mechanism driven by two linear actuators (P8-25-50-12-P, Actuonix Motion Devices, Inc.), providing two DoF for palmar flexion and dorsiflexion and for ulnar and radial deviation (Fig. ~\ref{fig:yui_hand}).

\subsubsection{Mobile platform}\label{sec314}
The mobile platform is a BLDC-motor-driven differential-drive base. It is customized from a commercially available differential-drive platform and includes four casters (two at the front and two at the rear).

In addition to planar mobility, the platform incorporates one active rotational DoF for knee flexion and extension. This allows vertical posture adjustment while maintaining stable upper-body motion during locomotion, reproducing torso movements characteristic of human walking. As shown in Fig.~\ref{fig:yui_structure}, the active knee flexion and extension mechanism is mechanically coupled with the passive hip and ankle joints, enabling the upper body to remain level during vertical motion. The knee DoF  is driven by the same motor unit as the waist, and a spring-based gravity compensation mechanism reduces the load on the knee actuator.

\subsection{Electrical system}\label{sec32}
\begin{figure}[t]
  \centering
  \includegraphics[width=1.0\textwidth]{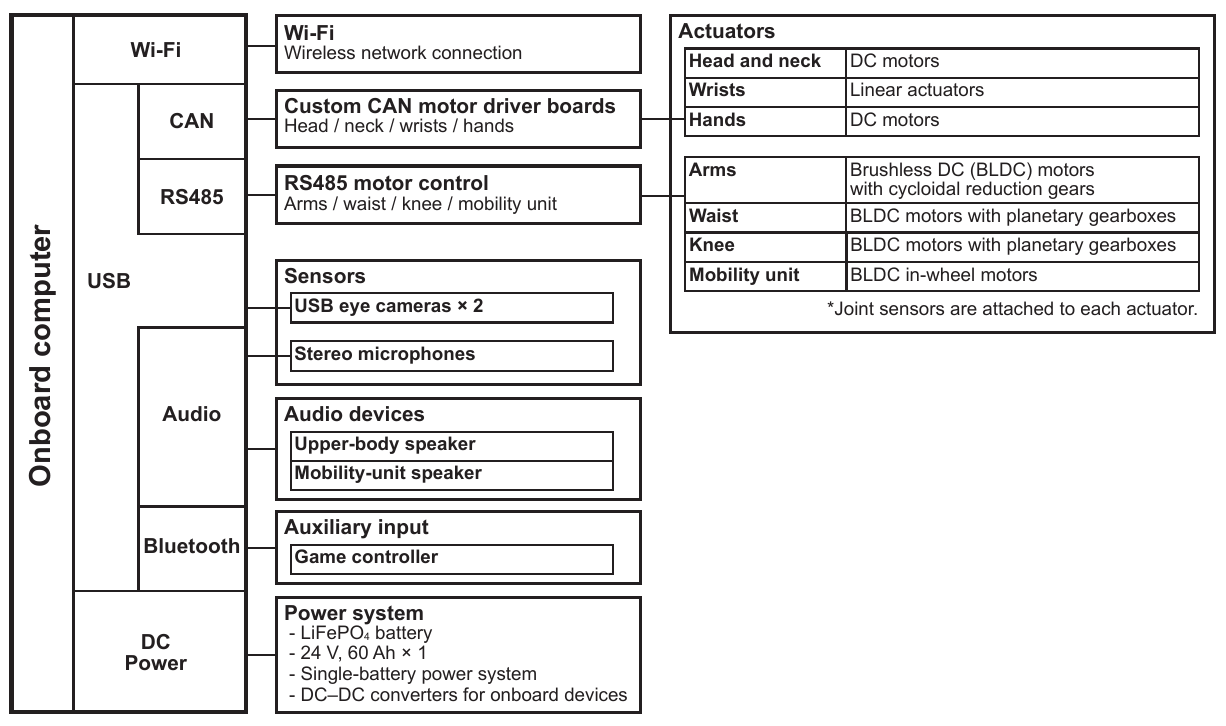}
  \caption{Electrical system overview.}
  \label{fig:yui_elecsys}
\end{figure}

In this subsection, we describe the electrical system configuration of Yui. The electrical system comprises a power supply system, motor control system, sensing system, audio devices, a network interface, and an auxiliary input device. Fig.~\ref{fig:yui_elecsys} summarizes the power distributions, control interfaces, sensor connections, and audio connections of the components. Each component is outlined in the following sections.

\subsubsection{Power supply system}\label{sec321}
The system uses a single onboard LiFePO$_4$ battery (\qty{24}{\volt}, \qty{60}{\ampere\hour}) mounted in the lower section of the mobile platform to maintain balanced weight distribution. Power from this battery is supplied to the onboard computer, motor drivers, actuators, sensors, and audio devices via dedicated power lines and DC-- converters. This single-battery design simplifies both charging and operation during long-term field deployment.
For deployment, we assumed extended operating hours at Expo 2025 (for example, during the pavilion opening hours from 10:00 to 21:00) and implemented an in-place charging scheme, allowing the battery to be charged without removal. To ensure continuous operation, two avatar units were prepared and used alternately (see Study~1 in Section~\ref{sec5} for details).

\subsubsection{Control boards}\label{sec322}
The main controller on the avatar side is a compact onboard PC installed within the body of Yui. This PC communicates with the operator-side PC through the teleoperation system described in Section~\ref{sec4} while managing the control interfaces for the actuators, sensors, and audio devices, as shown in Fig. ~\ref{fig:yui_elecsys}.

The actuators are controlled using two types of communication interfaces. The head, neck, wrists, and hands are driven using custom controller area network (CAN) motor driver boards connected to an onboard PC. The arms, waist, knee, and mobility units are controlled using RS485 communication. This division of the control interfaces reflects the hardware configuration of each subsystem while allowing the onboard PC to integrate the whole-body control.

\subsubsection{Sensing system}\label{sec323}
The sensing system consists of joint sensors, eye cameras, and stereo microphones. Joint sensors are attached to each actuator and are used to measure joint angles and linear actuator positions for position control and monitoring. Two USB cameras equipped with 200$^\circ$ wide-angle lenses are embedded in the eyes and provide visual feedback to the remote operator. Stereo microphones (CS-10EM, Roland Corporation) are embedded in the ears, which are used to capture audio from the interlocutor side.

\subsubsection{Audio device}\label{sec324}
To enable interaction with local interlocutors, Yui is equipped with speakers for audio playback of the operator's voice. A USB speakerphone (AT-CSP1, Audio-Technica Corporation) is mounted on the upper body, and an additional speaker (Sound Blaster GS3, Creative Technology Ltd.) is installed on the lower-rear section of the mobile platform to ensure adequate sound output in field environments. In the deployments and evaluations reported in this paper (Section~\ref{sec5}), the operator's voice was output through the speaker mounted on the mobile platform.

\section{Teleoperation System}\label{sec4}
\begin{figure}[t]
  \centering
  \includegraphics[width=\textwidth]{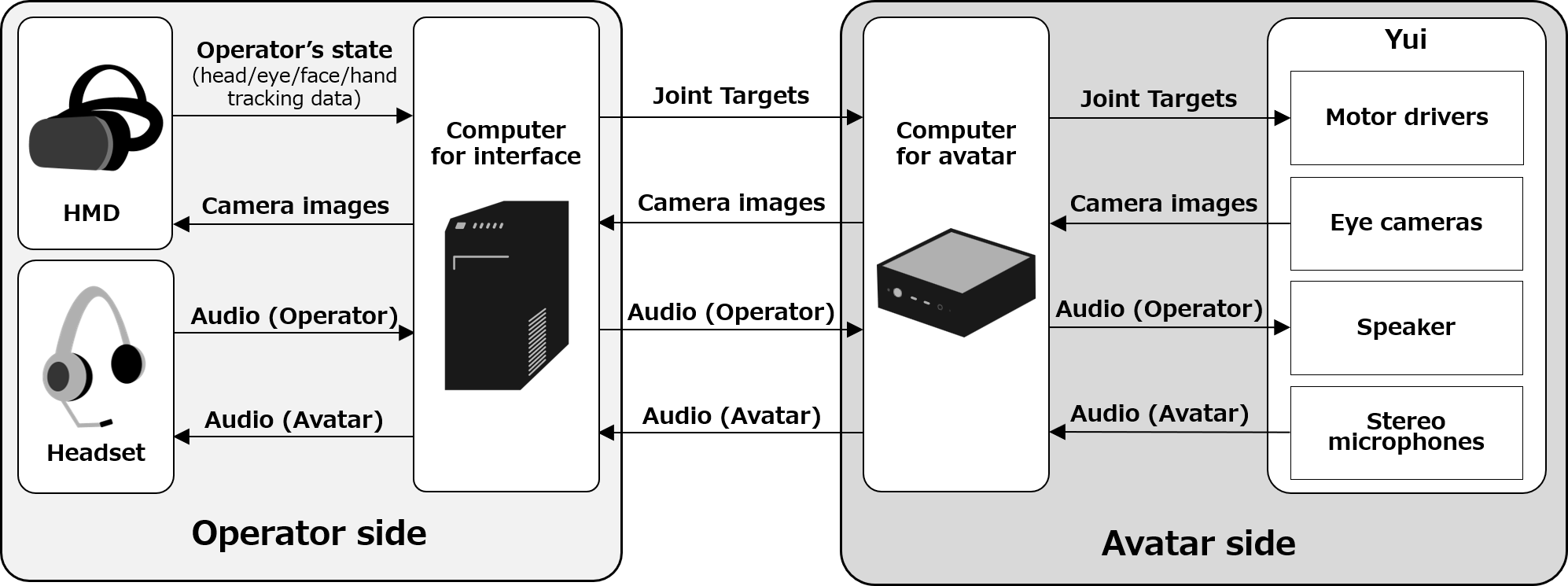}
  \caption{Data flow between the operator side and avatar side in the teleoperation system.}
  \label{fig:feedback_flow}
\end{figure}

In this section, we describe the teleoperation system used to remotely control the android avatar ``Yui''. 
Fig. \ref{fig:feedback_flow} shows an overview of the system.
Section~\ref{sec:hardware} introduces the hardware used by the operator. Section~\ref{sec:software} describes the software platform and communication, and Section~\ref{sec:data_flow} explains the detailed data flow and mapping logic.

\subsection{Operator interface and hardware} \label{sec:hardware}
In the immersive mode, the operator uses a commercially available HMD (Meta Quest Pro, Meta) as the control interface.  
The HMD supports both eye and face tracking.  
Although the HMD includes built-in speakers and a microphone, the operator wears an additional headset over the HMD to improve audio quality. A key feature of the system is that, by using the HMD's face, eye, and hand tracking together with pose tracking, the operator can control most of the avatar's body movements, except for locomotion, simply by wearing the HMD.
This enables intuitive operation without specialized training.
This ease of operation was a critical factor in the evaluation of general participants in Study~2 (Section~\ref{sec5}).  
The details of the mapping between the tracking data and avatar motion are described in Section~\ref{sec:o2a}.

The avatar's locomotion, including forward/backward movements and turning, is assigned to a joystick on a game controller and can be controlled visually. However, for safety reasons, the locomotion control from the operator side was disabled during the evaluation experiments reported in this paper. Therefore, the operator primarily controlled the avatar's facial expressions, gaze, head motion, arms, hands, and upper-body postures during evaluation.

In addition, a simplified control method without an HMD, referred to as the desktop mode, was developed.  
In this mode, only head movements, including facial expressions, are synchronized using a webcam.  
Facial expressions, eye movements, and head poses are tracked using MediaPipe~\cite{CLugaresi2019}, which is a real-time image-processing framework.  
The mapping between the tracking data and avatar motion in this mode is described in Section~\ref{sec:o2a}.  
The images captured by the avatar's eye cameras are displayed on a PC monitor, allowing the operator to wear a headset and interact with the avatar.

This mode enables remote interaction even for users who have difficulty wearing an HMD, such as children under 10 years of age, for whom the use of this HMD is restricted for health reasons.

\subsection{Software framework and networking} \label{sec:software}
A system based on Unity was constructed on both the operator and avatar sides, with communication established between the two Unity environments.
On the operator side, Unity was adopted to ensure seamless integration of data acquisition from the HMD and user interface rendering.
On the avatar side, a control system was implemented in Unity to receive the target values for each joint axis from the operator side and apply them as control inputs to the corresponding actuators.
Simultaneously, images captured by cameras embedded in the avatar's eyes and audio from its microphones are acquired and transmitted to the operator side in real-time, enabling bidirectional teleoperation and audiovisual communication.

Communication between the Unity environments on the operator and avatar sides is achieved using a WebRTC-based real-time software development kit (SDK) (Agora SDK, Agora, Inc.).
This SDK enables low-latency transmission of video and audio, along with data channel communication, allowing both media streaming and control commands to be handled within a single platform.
The end-to-end latency characteristics of the video, audio, and motion channels under the experimental teleoperation conditions used in this study are summarized in Appendix~\ref{sec:appendix_latency}.

\subsection{Data flow and mapping logic} \label{sec:data_flow}
The detailed data flow and mapping in the teleoperation of the avatar are described separately for the flows from the operator to the avatar and vice versa.

\subsubsection{Operator to avatar} \label{sec:o2a}
The data flow from the operator to the avatar is described by separating it into data mapping and processing on the avatar side.

\paragraph{Data Mapping}
In the immersive mode, the target values for facial expression; eye movement; head rotation; and hand, arm, and waist joints are calculated based on the tracking information obtained from the HMD, as described below. These values are transmitted to the avatar side via the data communication function of the real-time communication SDK.

For facial-expression synchronization, blendshape weights obtained from the HMD's face-tracking API are used, with values ranging from zero to one.
The HMD used in this study provides 63 blendshape weights based on a facial action coding system (FACS) \cite{Ekman1978}.
Based on these facial parameters, the target values for each facial actuation motor, also ranging from zero to one, are determined.
The mapping was manually adjusted so that the operator's facial expressions could be reproduced on the avatar as faithfully as possible.

For eye movements, the rotation angles obtained from the HMD's eye-tracking API are used as target values.
As the vertical movement of the avatar's eyes is mechanically linked between the left and right eyes, the average of the vertical rotation angles of both eyes is used as the target value.

The head rotation is computed from the roll, pitch, and yaw angles obtained from the HMD's pose tracking, and target values for each axis are calculated to reproduce the corresponding orientation.

For hand motion, the degree of flexion of each finger is calculated from the HMD's hand-tracking information as a value between zero and one, where zero represents a fully extended state and one represents maximum flexion, which are used as the target values for each finger.

For the arms and waist, the positions of the head and wrists are obtained from the HMD tracking data, and the joint angles are calculated using inverse kinematics.
For operational stability, only the yaw axis of the waist is controlled.

In the desktop mode, the target values for facial-expression motors and eye-control motors, ranging from zero to one, are determined based on 52 types of blendshape weights obtained from webcam images using MediaPipe.
The mapping is manually adjusted such that the operator's facial expressions are reproduced on the avatar as faithfully as possible.
The head rotation is calculated using facial orientation information obtained from MediaPipe, and target values for each axis are computed to reproduce the corresponding orientation.
The hand, arm, and waist motions are not synchronized in this mode and remain fixed in their initial states.

\paragraph{Data processing on the avatar side}
The target values for each joint received from the operator side are provided as inputs to the control system of each corresponding body part. 
The custom CAN motor driver boards for the head, neck, wrists, and hands are connected to the onboard PC via a USB-to-CAN adapter (PCAN-USB, PEAK-System), and control commands are sent from Unity C\# scripts using the official API. 
By contrast, the arms, waist, knee, and mobility unit are controlled via RS485 communication. 
The target values received by Unity are transmitted to the corresponding motor-control programs via socket communication, and the motor control is executed within each program.

In addition, the audio from the operator's headset microphone is received and played back through the speaker mounted on the avatar.
In the experiments conducted in this study, the operator's voice was processed using real-time voice conversion to maintain consistency between the avatar's appearance and vocal impression. The processed voice was output as a slightly higher and more avatar-appropriate voice while preserving the characteristics of the original speaker's voice.

\subsubsection{Avatar to operator}
The data flow from the avatar to the operator primarily consists of visual and auditory information.
Video from cameras embedded in the avatar's eyes and audio from stereo microphones in the ears are captured and transmitted to the operator using the real-time communication SDK's streaming functions. 
In the experiments reported in this study, only the video from a single-eye camera was provided to the operator to maintain communication stability.
Although the teleoperation system can be integrated with our previously proposed vision-sharing approach \cite{Shinkawa2024}, the field evaluations in this paper used this simplified visual-feedback configuration to prioritize stable operation in public settings with first-time general users.

In the immersive mode, the video received on the operator side is projected onto a virtual hemispherical screen and presented through the HMD.
The received audio is reproduced as spatial audio through the headset worn by the operator.
This configuration allows the operator to perceive audiovisual information from the remote environment in an immersive manner.

In the desktop mode, the received video is displayed on a monitor.

\section{Field Deployment and Evaluation Methods}\label{sec5}
In this section, we describe in detail the methodologies of the three studies (Studies 1--3) conducted to deploy and evaluate the developed android avatar, Yui, in real-world settings.
Study~1 examined the long-term in-the-wild deployment at Expo 2025 in Osaka, Kansai, Japan.
Study~2 focused on a special class and teleoperation experience conducted among elementary school students in Tokyo and Ishikawa.
Study~3 presented an evaluation of teleoperation and interaction among general participants in Ishikawa.
For each study, the procedures and evaluation methods are described in the following sections.

\subsection{Study 1: Long-Term ``In-the-Wild'' Deployment (Expo 2025 Osaka, Kansai, Japan)}
The objective of Study 1 was to examine the system's technical robustness in a public setting, including its stability and durability and qualitatively observe visitors' natural reactions to and social acceptance of the system.

\begin{figure}[t]
 \centering
    \begin{tabular}{cc}
        \begin{minipage}{0.5\textwidth}
            \centering                    
            \begin{subfigure}[t]{\columnwidth}
                \centering
                \includegraphics[height=6.5cm]{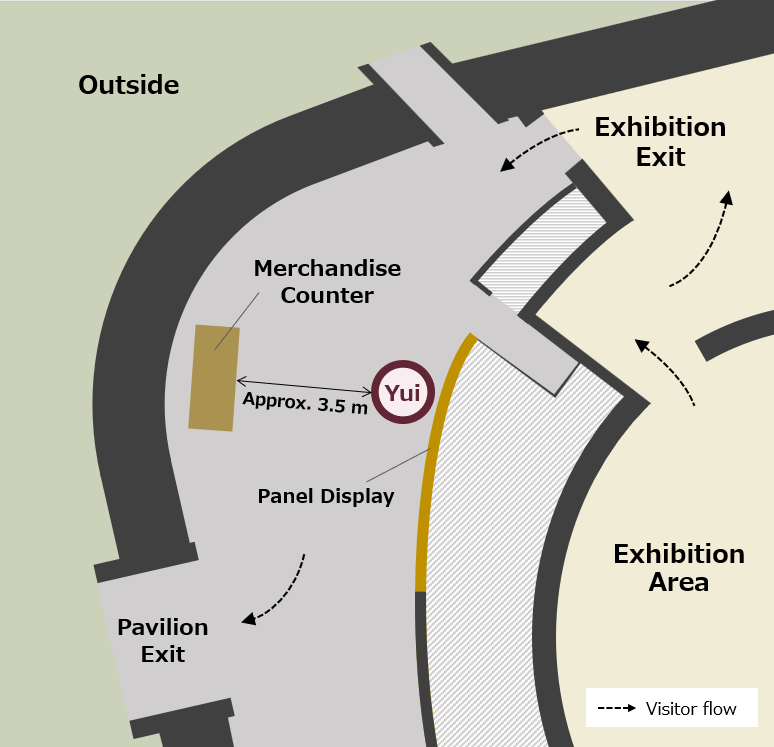}
                \subcaption{Location of Yui in the pavilion\\}
                \label{fig:expo_layout}
            \end{subfigure}
        \end{minipage}
        &
        \begin{minipage}{0.5\textwidth}
            \centering 
            \begin{subfigure}[t]{\columnwidth}
                \centering
                \includegraphics[height=6.5cm]{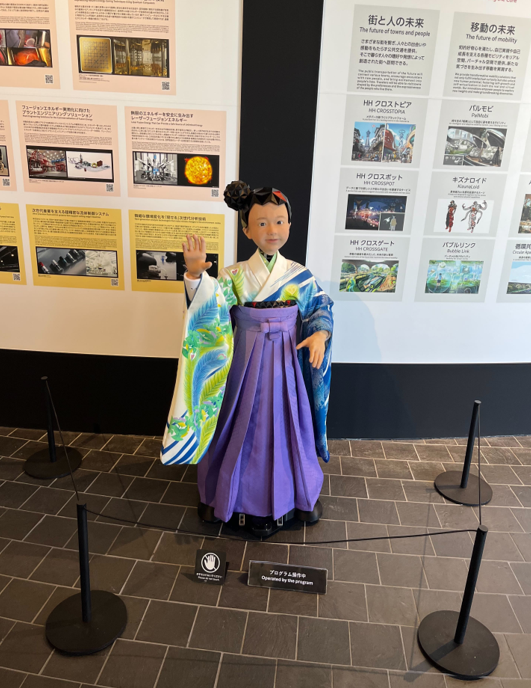}
                \subcaption{Photograph of Yui during the exhibition.}
                \label{fig:expo_photo}
            \end{subfigure}
        \end{minipage}
    \end{tabular}
    \caption{Exhibition of Yui in the pavilion during the Expo (Signature Pavilion: Future of Life).}
    \label{fig:expo_exhibition}
\end{figure}

The avatar was exhibited for approximately six months from April to October 2025 in the Pavilion of the Future of Life at Expo 2025 in Osaka, Kansai~\cite{FutureOfLife}.  
The installation locations within the pavilion are shown in Fig.~\ref{fig:expo_layout}.
Within the pavilion, a merchandise counter and display panel were located near the exit of the exhibition area, and the avatar was placed in front of the display panel.
None of the pavilion entrances and exits were equipped with doors, and the exhibition area was located in a semi-open environment with direct outdoor access.
As shown in Fig.~\ref{fig:expo_photo}, the avatar is enclosed by rope partitions during the exhibition.

To support long-term operations, two avatars were prepared, one of which was selected for display depending on the situation.
The details of the avatar's appearance are provided in Appendix~\ref{sec:appendix_avatar}.

Although the Yui system is designed for teleoperation, maintaining its continuous remote operation under the deployment conditions of approximately 11 h of daily exhibition over a period of six months was difficult.
Therefore, trained researchers who were proficient in remotely operating the avatar controlled Yui in advance, performing speech and actions corresponding to the exhibition content within the pavilion, and these were recorded.
A method in which the recorded data were repeatedly played back was defined as the autonomous playback mode, which was adopted as the primary operation mode.
By contrast, the remote operation mode was employed in specific situations, such as when responding to special visitors or media coverage or when an operator was available.

The remote operation of the avatar was conducted by researchers either from the University of Electro-Communications or from an operator space within the pavilion using the WebRTC-based system described in Sections~\ref{sec2} and \ref{sec4}.
Due to ethical and operational constraints, structured questionnaire data were not collected.

In this study, the technical performance metrics of the system included the total operating time, average daily operating time, and records of failures or system interruptions that occurred during the deployment, along with their causes and countermeasures.
Additionally, qualitative data on social acceptance were collected by observing and recording visitors' behaviors, and their interactions with the avatar during the exhibition period, with photographs and videos obtained with permission.
These results are reported in Section~\ref{sec:result_study1}.

\subsection{Study 2: Field Report on Remote Educational Exchange for Elementary School Students}
The objective of Study 2 was to examine the potential of remote interaction mediated by an android avatar and observe how interactions between elementary school students are established through the avatar.

In this study, third-grade elementary school students in Chofu, Tokyo, aged 8--9 years, acted as operators and engaged in remote interaction via the avatar with third- and fourth-grade students in Kanazawa, Ishikawa, aged 8--10 years, located approximately \qty{280}{km} away.
The avatar was placed on the Ishikawa side and operated remotely by the students in Tokyo.
The event was conducted during a special class.

\begin{figure}[t]
  \centering
  \includegraphics[width=0.5\textwidth]{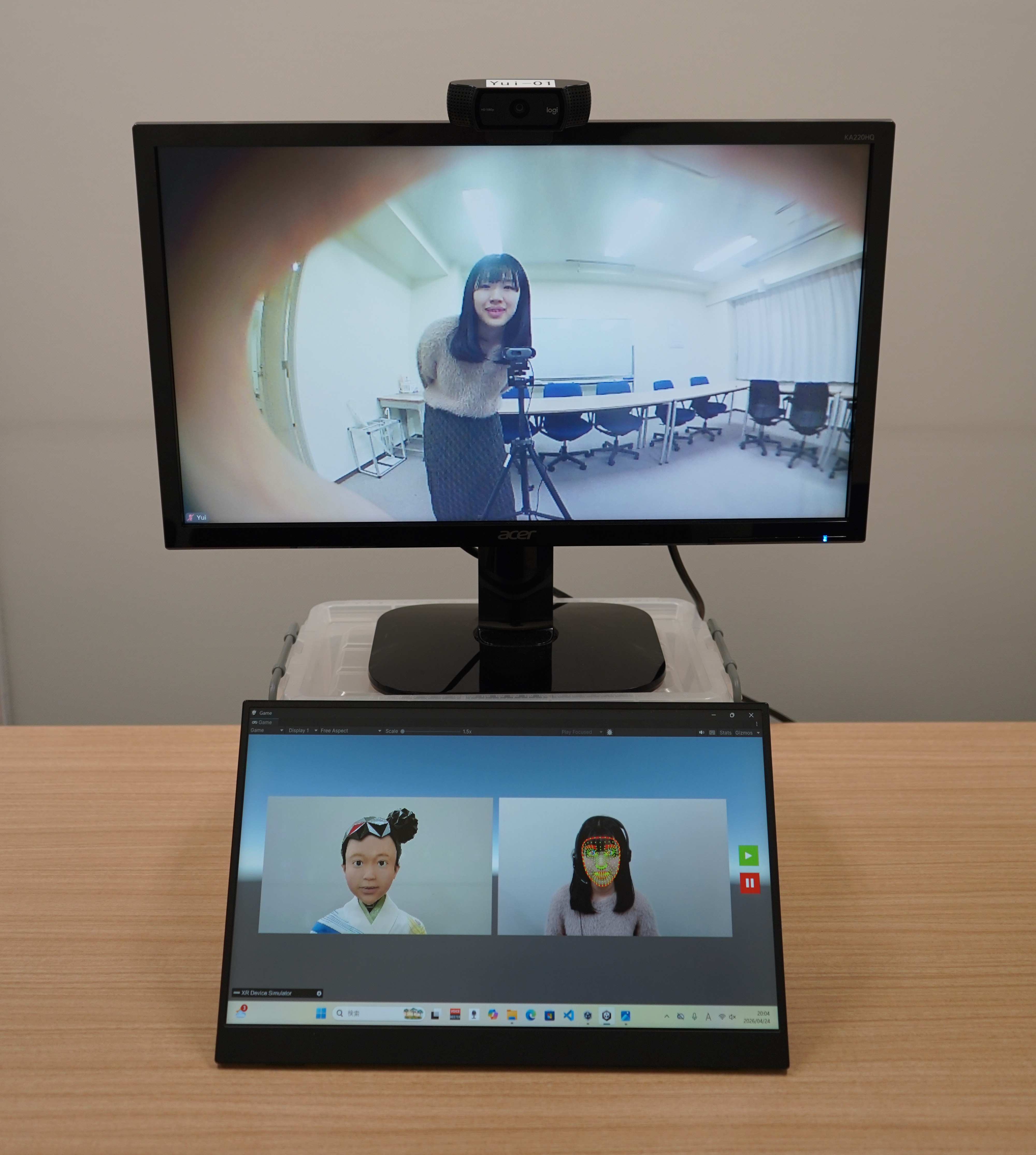}
  \caption{Operator-side monitor displays in Study 2. The photograph was recreated by one of the authors.}
  \label{fig:school_monitors}
\end{figure}

On the operator side, the desktop mode was used to control the avatar's head.
Students in Tokyo who wished to operate the avatar wore a headset and engaged in one-on-one conversations with students on the Ishikawa side via the avatar while watching the monitor.
In this event, two monitors were prepared, showing not only the Yui-perspective view but also the operator's face together with Yui's face (see Fig. ~\ref{fig:school_monitors}). The aim was to recognize that the students were remotely controlling the avatar's head.

Because this activity was conducted in collaboration with an educational setting, no structured questionnaires or quantitative data were collected because of ethical or operational constraints.
Instead, the establishment of remote interaction and operational challenges were examined based on observational records during the class and feedback from teachers.
These results are reported in Section~\ref{sec:result_study2}.

\subsection{Study 3: Public Operator-Interlocutor Evaluation and Interlocutor Impression Evaluation (Ishikawa)}
The objective of Study 3 was to evaluate the experiences and impressions of the general participants when operating and interacting with Yui.
This evaluation, conducted in Kanazawa, Ishikawa, consisted of the following two experiments.

\begin{itemize}
\item {\bf Experiment 1 (Operator--Interlocutor Pairs)}: General participants were assigned to pairs consisting of an operator and an interlocutor, and they interacted via the avatar.
\item {\bf Experiment 2 (Interlocutor Impression)}: General participants interacted with the avatar operated by a trained operator.
\end{itemize}

The avatar operation was performed in the immersive mode in both the experiments.
All the experimental participants were limited to individuals aged 18 years or older.
For event participants under the age of 18, a remote operation experience using desktop mode was provided.

\subsubsection{Participants}
The participants were recruited from visitors attending an event held in Kanazawa, Ishikawa.

In Experiment 1, 118 participants, including 62 operators and 56 interlocutors, responded to the questionnaire.
After excluding two participants who did not complete the questionnaire and 15 participants whose data were affected by avatar malfunctions during the interaction, the data from 108 participants, including 52 operators and 49 interlocutors, were used for analysis.
The age distribution of the participants is shown in Fig. \ref{fig:dist_exp1} in Appendix \ref{sec:appendix_dist}.

In Experiment 2, 17 participants responded to the questionnaire.  
After excluding one participant who did not complete the questionnaire, the data from 16 participants, including 6 males and 10 females, were used for analysis.
The age distribution of the participants is shown in Fig. \ref{fig:dist_exp2} in Appendix \ref{sec:appendix_dist}.

The event was a special exhibition that included a teleoperation experience of the android avatar, Yui, which had been exhibited at the Expo, as well as a hands-on experience of Kaga Yuzen, a traditional craft of the region, and demonstrations of the latest technologies by NTT DOCOMO, which was an  event organizer.
To recruit participants, the event was announced in advance through the official websites of the organizers.

When recruiting participants, we clarified that event attendees might be invited to participate voluntarily in the experiment.

\subsubsection{Materials}
The avatar used in this study was the android avatar, Yui, described in Section~\ref{sec3}.
The operator used the HMD (Meta Quest Pro) described in Section~\ref{sec:hardware}.

In Experiments 1 and 2, questionnaires were administered after interactions via the avatar.
The questionnaires were voluntary, and participants were considered to have consented to participate in the experiment after completing the questionnaire.
The questionnaire items are listed in Table~\ref{tab:questions}.

Questions C-1, C-2, and C-3 were common to both operators and interlocutors, which evaluated the quality of the experience compared with video calls, sense of co-presence, and willingness to use the avatar, respectively.
Questions O-1, O-2, and O-3 were answered only by the operators and evaluated the degree to which intended operations were achieved, ease of understanding the operation method, and physical workload during the operation, respectively.
Questions I-1 and I-2 were answered only by the interlocutors, who evaluated the human-likeness of the avatar and the ease of conveying emotions and intentions, respectively.

Question O-3 was evaluated using a Visual Analog Scale ranging from 0 to 100, while the other items were rated on a seven-point Likert scale.  
The details of these scales are listed in Table~\ref{tab:scales}.
In Experiment 2, for Questions C-1 and C-3, participants were optionally asked to provide free-text explanations immediately after each question.

Prior to the questionnaire items listed in Table~\ref{tab:questions}, the participants' age group, gender, and relationship with the interlocutor in Experiment 1, and  frequency of video calls were collected.
The frequency of video calls was measured on a four-point scale consisting of almost never, a few times per month, once per week, and a few times per week.

Additionally, to ensure the stability of the avatar's facial expressions, the lower eyelids were fixed, and the facial expressions were controlled based on symmetrical target values.

\begin{table}[t]
\renewcommand{\arraystretch}{1.2}
\caption{Questionnaire items used in Experiments 1 and 2}
\centering
\begin{tabular}{m{7mm} m{55mm} cc c}
\hline
 & & \multicolumn{2}{c}{Experiment 1} & Experiment 2 \\
Item & Question & Operator & Interlocutor &  \\
\hline
C-1 & How did you find this remote interaction experience compared to a video call? 
& $\checkmark$ & $\checkmark$ & $\checkmark$ \\ \hdashline[2pt/1pt]
C-2 & Did you feel as if the other person was in the same physical space as you? 
& $\checkmark$ & $\checkmark$ & $\checkmark$ \\ \hline
O-1 & Were you able to control the robot avatar as intended? 
& $\checkmark$ & & \\ \hdashline[2pt/1pt]
O-2 & Was the operation intuitive and easy to understand? 
& $\checkmark$ & & \\ \hdashline[2pt/1pt]
O-3 & To what extent did you experience physical burden? 
& $\checkmark$ & & \\ \hline
I-1 & Did the robot avatar's responses and behaviors feel human-like? 
& & $\checkmark$ & $\checkmark$ \\ \hdashline[2pt/1pt]
I-2 & Did the other person's emotions and intentions feel well conveyed through the robot avatar?
& & $\checkmark$ & $\checkmark$ \\ \hdashline[2pt/1pt]
C-3 & Would you like to use this type of robot-avatar-based remote communication tool? 
& $\checkmark$ & $\checkmark$ & $\checkmark$ \\ \hline
\end{tabular}
\label{tab:questions}
\end{table}

\begin{table}[t]
\caption{Response scales used in the questionnaire}
\renewcommand{\arraystretch}{1.2}
\centering
\begin{tabular}{m{30mm} m{90mm}}
\hline
Item & Response Scale \\
\hline
C-1 & 7-point scale: Very bad, Bad, Slightly bad, Neutral, Slightly good, Good, Very good \\ \hdashline[2pt/1pt]
C-2, O-1, O-2, I-1, I-2 & 
7-point scale: Strongly did not feel so, Did not feel so, Slightly did not feel so, Neutral, Slightly felt so, Felt so, Strongly felt so \\ \hdashline[2pt/1pt]
C-3 & 
7-point scale: Strongly disagree, Disagree, Slightly disagree, Neither agree nor disagree, Slightly agree, Agree, Strongly agree \\ \hdashline[2pt/1pt]
O-3 & 
Visual Analog Scale (VAS): 0--100 \\
\hline
\end{tabular}
\label{tab:scales}
\end{table}

\subsubsection{Procedure}
Experiments 1 and 2 were conducted as follows.

\paragraph{Experiment 1: Operator--interlocutor pairs}
This experiment was conducted as part of an experiential event.
Prior to the experiment, the participants were informed that the purpose of the study was to investigate the robot avatar operation and impressions of both operators and interlocutors. 
They were also informed that they would be assigned to the roles of operator and interlocutor and would engage in an approximately two-minute interaction via the avatar.

After the explanation, the participants were instructed to form pairs consisting of an operator and interlocutor.
The interlocutor was instructed to move to a separate room where the avatar was located and to stand in front of the avatar.

The operator was instructed to wear the HMD and headset.
After synchronization with the avatar, we confirmed that the motion synchronization and audio communication were functioning properly.  
Subsequently, a free interaction between the operator and interlocutor via the avatar was conducted.
The interaction lasted approximately two minutes and was terminated by the experimenter within one minute after the allotted time had elapsed.

Immediately after the interaction, the participants were instructed to voluntarily complete the questionnaire.
At the beginning of the questionnaire, an explanatory statement describing the purpose of the study, data handling, voluntary nature of participation, right to withdraw at any time, and planned publication of the results was provided.
The participants provided informed consent by confirming the statement and selecting the consent option.

During the operation, the avatar's facial expressions, eye movements, head rotation, arm movements, waist rotation, and hand movements were synchronized with those of the operator.
The arm movements were constrained to avoid excessive loads on the motors and prevent the avatar from contacting its own body.  
Although the hand movements were also synchronized, the range of finger motion was limited owing to the condition of the hand mechanism during the experiment; therefore, this limitation was not explicitly explained to the participants.
The interlocutor was instructed to stand at a distance of approximately \qty{1}{m} from the avatar, and care was taken to avoid physical contact as much as possible.

For participants who played multiple roles, the questionnaire responses were collected after only their first role.

\paragraph{Experiment 2: Interlocutor impression evaluation}
As in Experiment 1, this experiment was conducted as part of an experiential event.
The participants assumed the role of interlocutors and engaged in free interaction with an avatar remotely operated by a trained researcher.
The operator controlled the avatar from a separate room.

After the interaction, the participants were asked to voluntarily complete a questionnaire.
As in Experiment 1, an explanatory statement regarding data handling and other relevant information was presented at the beginning of the questionnaire, and participants provided informed consent by confirming the statement and selecting the consent option.
Participants in Experiment 1 were excluded from this experiment.

\subsubsection{Ethical considerations}
The study protocol was approved by the Ethics Committee of the University of Electro-Communications (No. H25054).
All the participants provided informed consent through a web-based form before participating in the experiment.

\section{Results and Implications}
The results and implications of these three studies are the following.

\subsection{Study 1: Long-Term ``In-the-Wild'' Deployment (Expo 2025 Osaka, Kansai, Japan)}\label{sec:result_study1}
The results and implications of Study~1 are the following.

\subsubsection{Results}
The avatar was continuously exhibited for approximately six months, with a total operating time of approximately 1131 h across the two units.  
The decision to operate the exhibition on a given day was made by the research team based on the condition of the avatar and exhibition environment; in some cases, the exhibition was suspended for an entire day because of rain or system malfunctions.  
The exhibition lasted 113 days, with an average daily operating time of approximately 10 h.

\begin{figure}[t]
  \centering
  \includegraphics[width=0.8\textwidth]{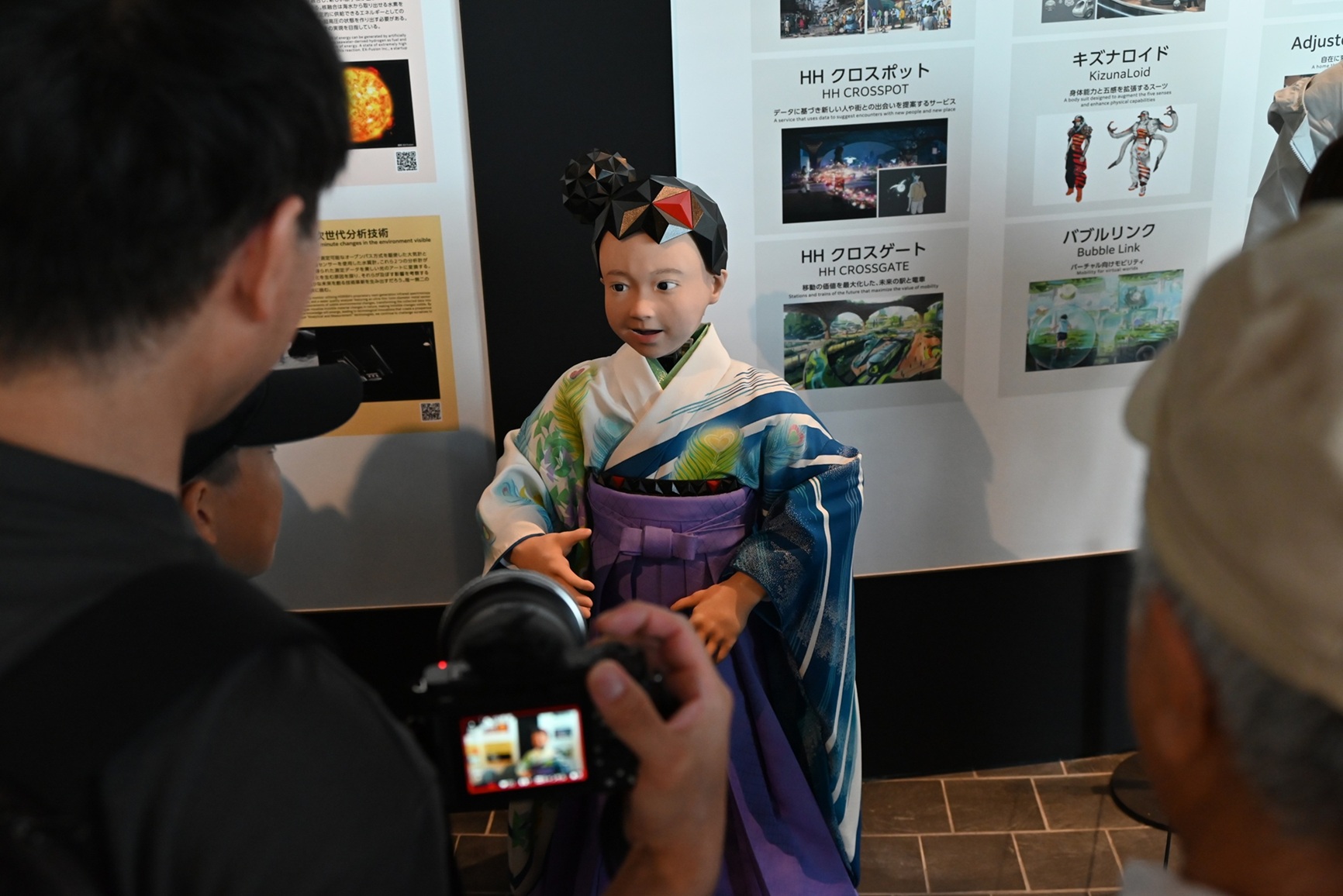}
  \caption{Yui interacting with visitors at Expo 2025 Osaka, Kansai, Japan (Signature Pavilion: Future of Life).}
  \label{fig:expo_yui}
\end{figure}

During the exhibition period, the avatar performed farewell interactions with visitors using either the autonomous playback mode or teleoperation using the immersive mode.  
An example of the avatar interacting with visitors during teleoperation is shown in Fig. ~\ref{fig:expo_yui}.

Several technical issues related to motors and mechanical components occurred during the exhibition period; however, the operation was maintained by addressing these issues through component replacement and software adjustments, as needed.  
Only when critical failures occurred that made continued operation difficult was the exhibition suspended for several days until repairs were completed.  
Near the end of the exhibition period, battery degradation reduced the continuous operating time, making full-day battery power operation difficult.  
Therefore, approximately two weeks before the end of the exhibition, the system was switched to operation with an external power supply.

\subsubsection{Implications}
These results suggest that the proposed system can be operated for an extended period in a public environment.
Although several technical issues occurred during the exhibition, the system maintained a continuous operation through appropriate maintenance and adjustments, suggesting that it had a level of robustness suitable for real-world deployment.

Furthermore, during the exhibition period, visitors spontaneously approached the avatar, responded to its speech and gestures, and captured photographs in close proximity.
In addition, during teleoperation using the immersive mode, visitors frequently engaged in conversations with the remote operator through the avatar.
These observations suggest that the system may be accepted by diverse visitors in public spaces and may elicit interactive engagement.

Thus, the long-term deployment at Expo 2025 broadly supported the objective of Study 1, namely, examining the technical robustness of the system and its social acceptance in a public environment.

\subsection{Study 2: Field Report on Remote Educational Exchange for Elementary School Students}\label{sec:result_study2}
The results and implications of Study~2 are the following.

\subsubsection{Results}
\begin{figure}[t]
 \centering
    \begin{tabular}{cc}
        \begin{minipage}{0.4\textwidth}
            \centering                    
            \begin{subfigure}[t]{\columnwidth}
                \centering
                \includegraphics[height=4.1cm]{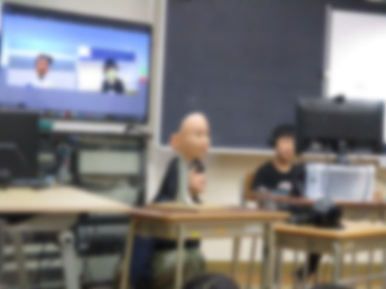}
                \subcaption{Tokyo side}
                \label{fig:school_chofu}
            \end{subfigure}
        \end{minipage}
        &
        \begin{minipage}{0.6\textwidth}
            \centering 
            \begin{subfigure}[t]{\columnwidth}
                \centering
                \includegraphics[height=4.1cm]{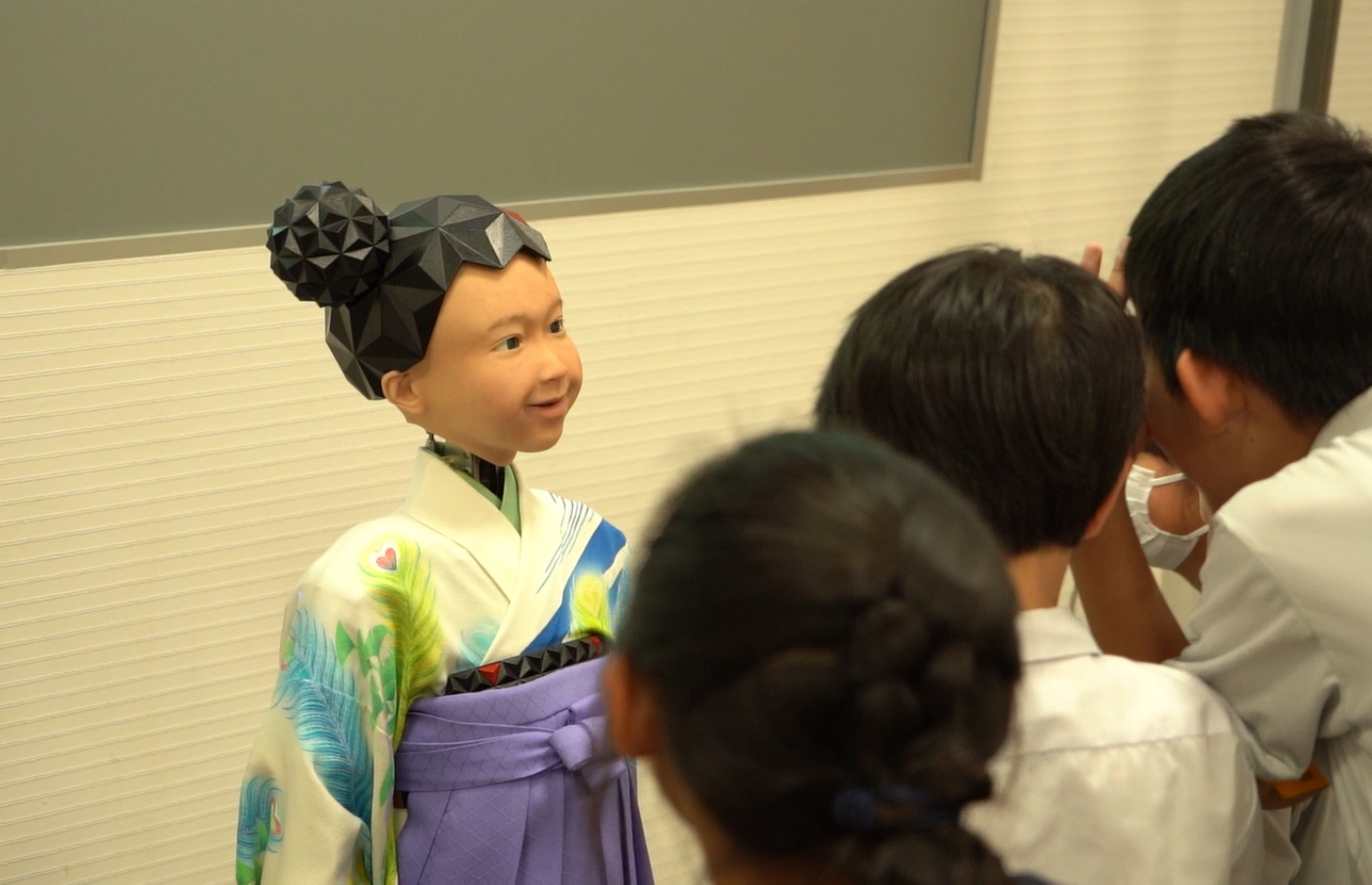}
                \subcaption{Ishikawa side}
                \label{fig:school_kanazawa}
            \end{subfigure}
        \end{minipage}
    \end{tabular}
    \caption{Remote avatar interaction among elementary school students in Chofu, Tokyo and Kanazawa, Ishikawa.
In the Tokyo side, the object placed on the front desk was a prototype android head presented only as a visual reference for the children and was not a part of the teleoperation system used in this study.}
    \label{fig:school_event}
\end{figure}

Fig.~\ref{fig:school_chofu} shows elementary school students in Chofu, Tokyo, remotely operating the avatar, whereas Fig. ~\ref{fig:school_kanazawa} shows students in Kanazawa, Ishikawa, interacting face-to-face with the avatar.

In the first half of the event, the locations of the participating students and role of the avatar were explained in a lecture-style format.
Subsequently, a small number of students from Tokyo and Ishikawa, who had been selected in advance, engaged in one-to-one interactions via the avatar.
Students in Tokyo wore headsets and operated the avatar while receiving instructions on its operation, including controlling facial expressions.

The students interacted in a question-and-answer format and discussed topics such as self-introduction, personal interests, and daily school life, thereby facilitating mutual communication.

After the event, feedback collected by the teacher in charge on the Tokyo side included comments such as ``It is amazing that Yui can move on my behalf,'' ``I was surprised that there is a robot that allows us to talk even when we are far apart,'' and ``It is impressive that the robot can be controlled using a camera.''

\subsubsection{Implications}
The use of the desktop mode was readily accepted by children aged 8--9 years, as it did not require wearing an HMD.
The feedback from the participating children indicated a strong interest in the concept of a robot avatar that enables communication with people in remote locations.

However, because the desktop mode is limited to head synchronization, situations existed in which the operator's body movements were not sufficiently reflected in the avatar.
Consequently, some children who operated on the avatar experienced confusion.
In addition, because the visual feedback was presented on a display, the level of immersion was limited compared with the immersive mode, suggesting a remaining challenge in terms of immersive experience.

Nevertheless, through this activity, the children attempted to communicate via the avatar and engage in mutual interactions.
They also showed a high level of interest in this novel form of communication, in which they interacted with distant peers through a robot avatar.
These observations suggest that remote interactions mediated by android avatars may facilitate communication between children.

Thus, Study 2 provided exploratory evidence suggesting that remote interaction between geographically separated elementary school students can be established through the android avatar in an educational setting.

\subsection{Study 3: Public Operator--Interlocutor Evaluation and Interlocutor Impression Evaluation (Ishikawa)}
The results of Experiments 1 and 2 are the following.

\subsubsection{Experiment 1: Operator--interlocutor pairs}
The results and implications of Experiment 1 are the following.

\begin{figure}[tp]                 
    \begin{subfigure}[t]{\columnwidth}
        \centering
        \includegraphics[width=0.9\columnwidth]{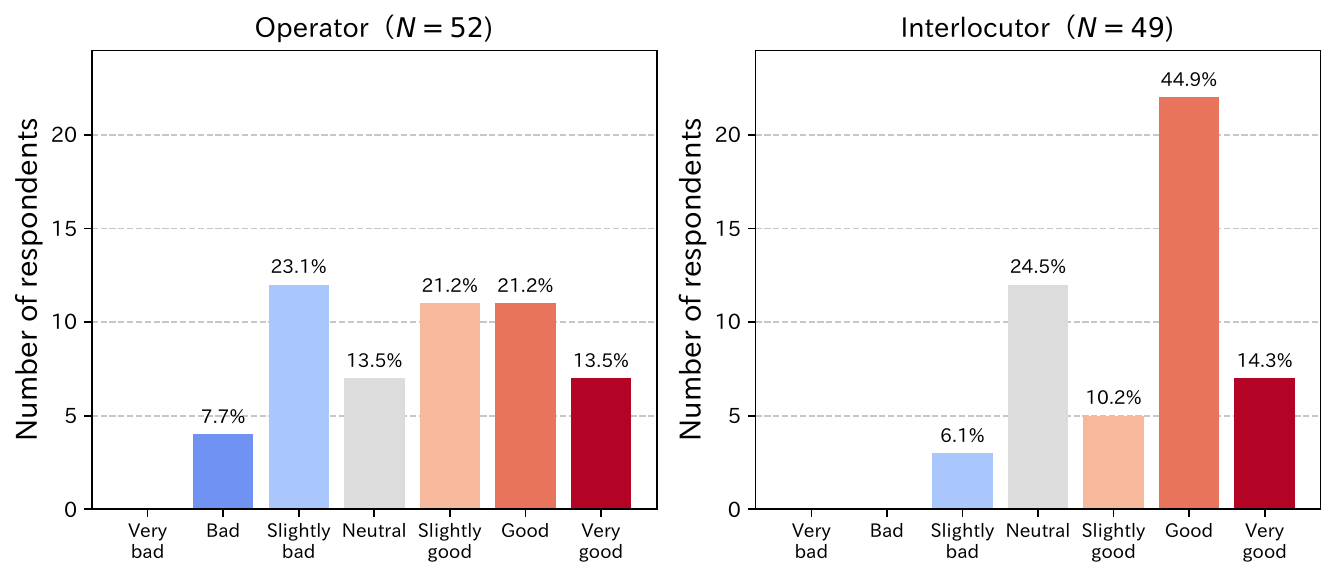}
        \subcaption{Question C-1: How did you find this remote interaction experience compared to a video call?}
        \label{fig:question_c1}
    \end{subfigure}
    \\
    \begin{subfigure}[t]{\columnwidth}
        \centering
        \includegraphics[width=0.9\columnwidth]{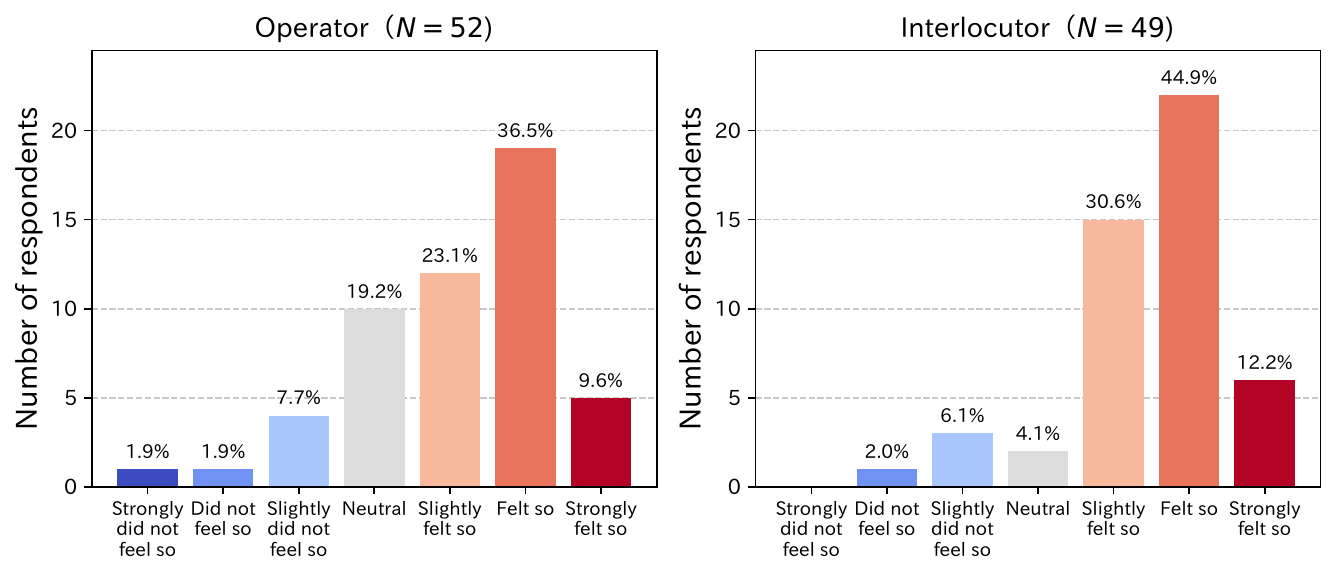}
        \subcaption{Question C-2: Did you feel as if the other person was in the same physical space as you?}
        \label{fig:question_c2}
    \end{subfigure}
    \\
    \begin{subfigure}[t]{\columnwidth}
        \centering
        \includegraphics[width=0.9\columnwidth]{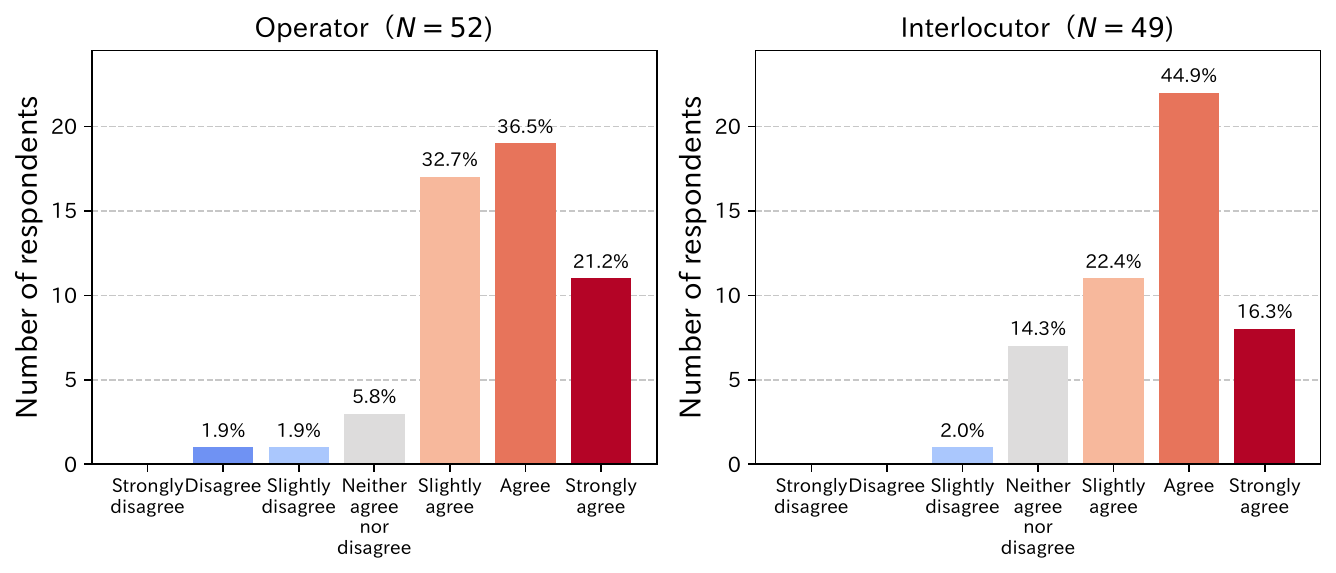}
        \subcaption{Question C-3: Would you like to use this type of robot-avatar-based remote communication tool?}
        \label{fig:question_c3}
    \end{subfigure}
    \caption{Distribution of responses to common questions in Experiment 1.}
    \label{fig:common_questions}
\end{figure}

\begin{figure}[t]
 \centering
    \begin{tabular}{cc}
        \begin{minipage}{0.5\textwidth}
            \centering                    
            \begin{subfigure}[t]{\columnwidth}
                \centering
                \includegraphics[height=5cm]{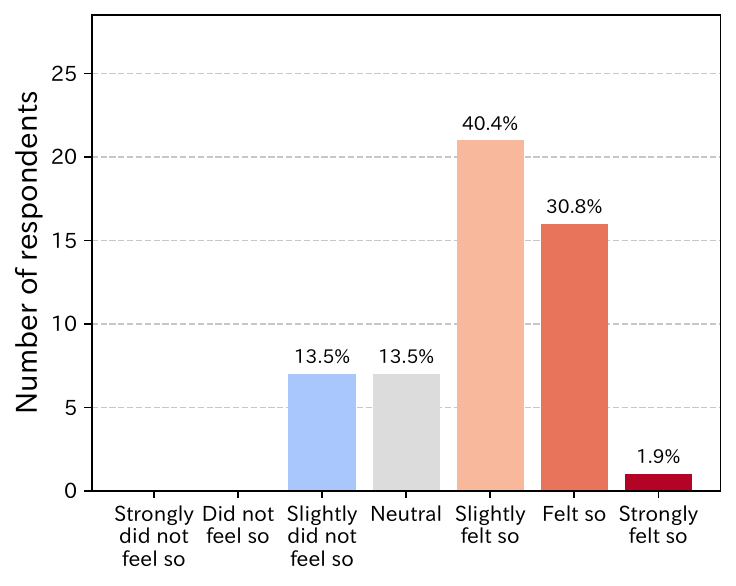}
                \subcaption{Question O-1: Were you able to control the robot avatar as intended?}
                \label{fig:question_o1}
            \end{subfigure}
        \end{minipage}
        &
        \begin{minipage}{0.5\textwidth}
            \centering 
            \begin{subfigure}[t]{\columnwidth}
                \centering
                \includegraphics[height=5cm]{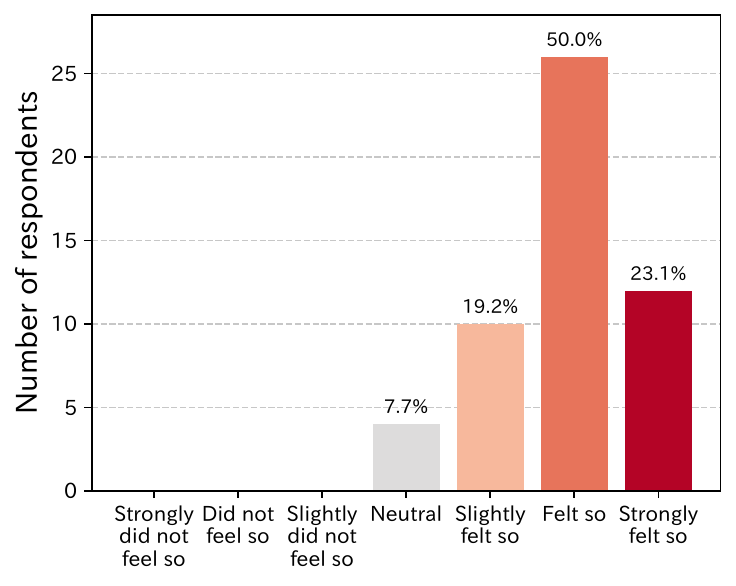}
                \subcaption{Question O-2: Was the operation intuitive and easy to understand?}
                \label{fig:question_o2}
            \end{subfigure}
        \end{minipage}
    \end{tabular}
    \\
    \begin{subfigure}[t]{\columnwidth}
        \centering
        \includegraphics[width=0.6\columnwidth]{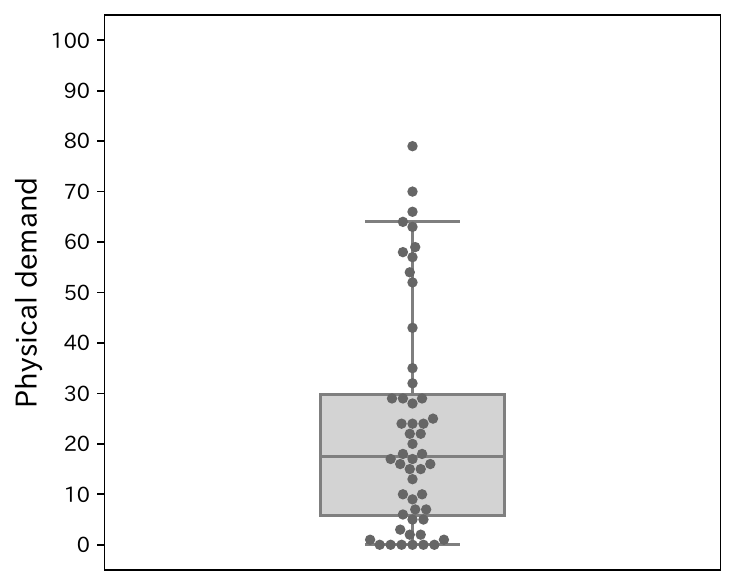}
        \subcaption{Question O-3: To what extent did you experience physical burden?}
        \label{fig:question_o3}
    \end{subfigure}
    \caption{Distribution of responses to operator-specific questions in Experiment 1 ($N=52$).}
    \label{fig:operator_questions}
\end{figure}

\begin{figure}[t]
 \centering
    \begin{tabular}{cc}
        \begin{minipage}{0.5\textwidth}
            \centering                    
            \begin{subfigure}[t]{\columnwidth}
                \centering
                \includegraphics[height=4.5cm]{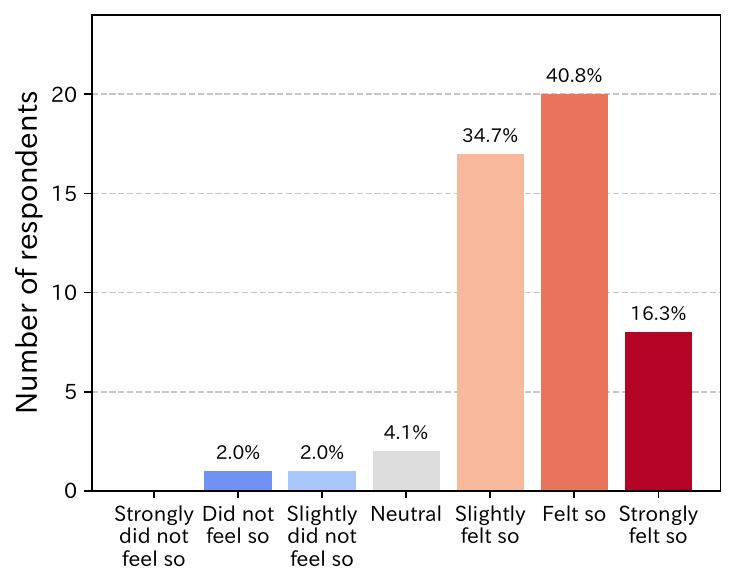}
                \subcaption{Question I-1: Did the robot avatar's responses and behaviors feel human-like?\\}
                \label{fig:question_i1}
            \end{subfigure}
        \end{minipage}
        &
        \begin{minipage}{0.5\textwidth}
            \centering 
            \begin{subfigure}[t]{\columnwidth}
                \centering
                \includegraphics[height=4.5cm]{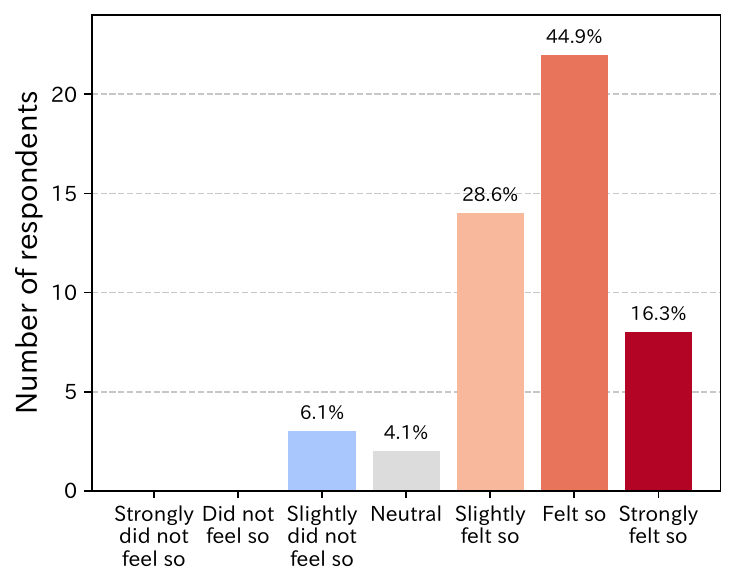}
                \subcaption{Question I-2: Did the other person's emotions and intentions feel well conveyed through the robot avatar?}
                \label{fig:question_i2}
            \end{subfigure}
        \end{minipage}
    \end{tabular}
    \caption{Distribution of responses to interlocutor-specific questions in Experiment 1 ($N=49$).}
    \label{fig:interlocutor_questions}
\end{figure}

\paragraph{Results}
The results for the common questions (C-1, C-2, and C-3), which were answered by both operators and interlocutors, are shown in Fig.~\ref{fig:question_c1}, Fig.~\ref{fig:question_c2}, and Fig.~\ref{fig:question_c3}, respectively.  
The results for the operator-specific questions (O-1, O-2, and O-3) are shown in Fig.~\ref{fig:question_o1}, Fig.~\ref{fig:question_o2}, and Fig.~\ref{fig:question_o3}, respectively. 
The results for the interlocutor-specific questions (I-1 and I-2) are shown in Fig.~\ref{fig:question_i1} and Fig.~\ref{fig:question_i2}, respectively.

\begin{figure}[t]
 \centering
    \begin{subfigure}[t]{\columnwidth}
        \centering
        \includegraphics[width=\columnwidth]{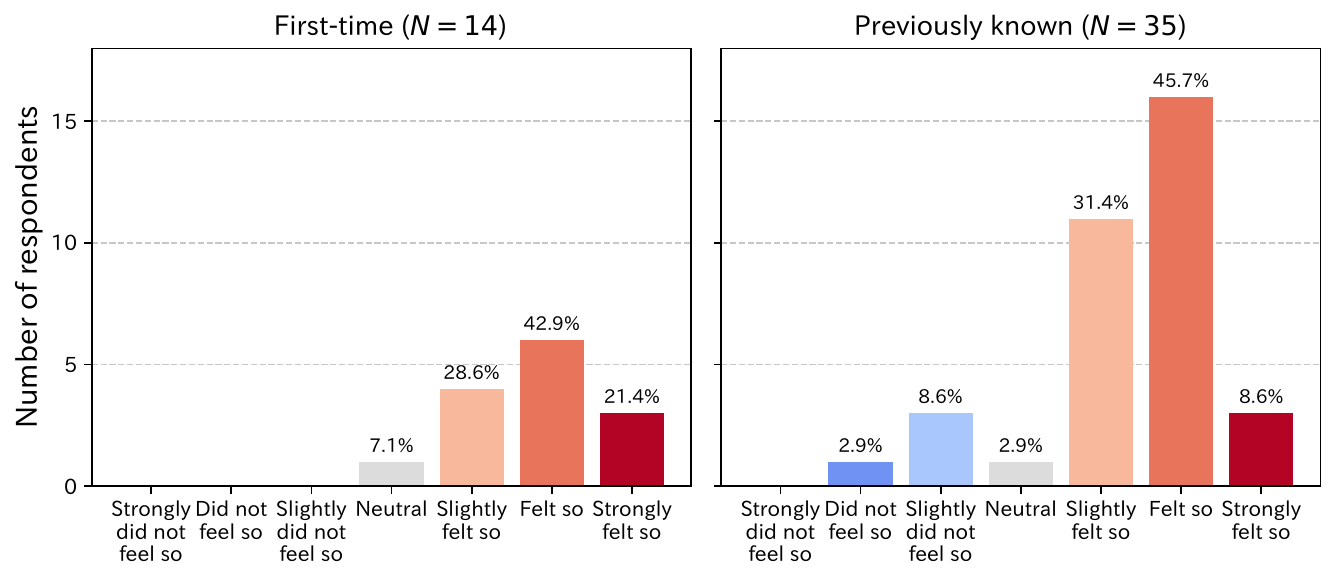}
        \subcaption{Question C-2: Did you feel as if the other person was in the same physical space as you?}
        \label{fig:known_c2}
    \end{subfigure}
    \\
    \begin{subfigure}[t]{\columnwidth}
        \centering
        \includegraphics[width=\columnwidth]{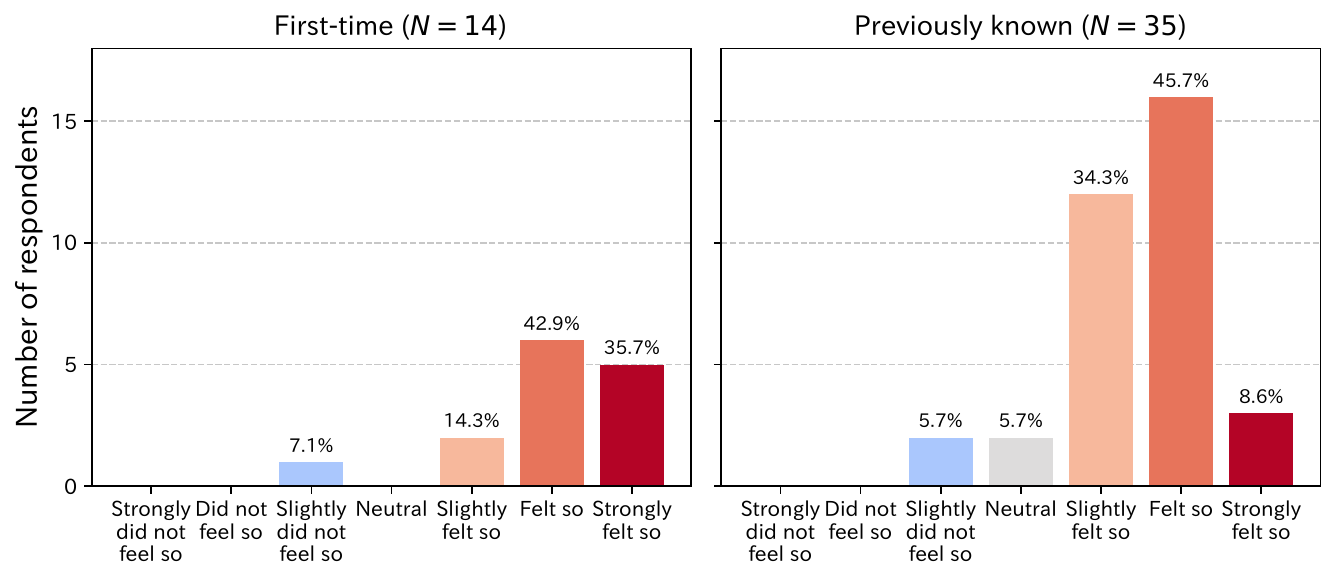}
        \subcaption{Question I-2: Did the other person's emotions and intentions feel well conveyed through the robot avatar?}
        \label{fig:known_i2}
    \end{subfigure}
    \caption{Distribution of responses to Questions C-2 and I-2 by familiarity with the interlocutor (first-time vs. previously known).}
    \label{fig:known_c2_i2}
\end{figure}

\begin{table}[t]
\renewcommand{\arraystretch}{1.2}
\centering
\caption{Proportions of positive and negative responses for each Likert-scale questionnaire item in Experiment 1}
\label{tab:exp1_results_summary}
\begin{tabular}{
>{\raggedright\arraybackslash}p{35mm}
>{\raggedright\arraybackslash}m{22mm}
>{\raggedleft\arraybackslash}m{20mm}
>{\raggedleft\arraybackslash}m{20mm}
}
\toprule
Question & Role 
& \multicolumn{1}{>{\raggedright\arraybackslash}m{20mm}}{Positive response$^{\dagger}$ (\%)} 
& \multicolumn{1}{>{\raggedright\arraybackslash}m{20mm}}{Negative response$^{\dagger}$ (\%)} \\
\midrule

\multirow[c]{2}{35mm}{C-1 (Comparison with video calls)}
    & Operator     & 55.8 & 30.8 \\
    & Interlocutor & 69.4 & 6.1  \\
\hdashline[2pt/1pt]

\multirow[c]{2}{35mm}{C-2 (Sense of co-presence)}
    & Operator     & 69.2 & 11.5 \\
    & Interlocutor & 87.8 & 8.2  \\
\hdashline[2pt/1pt]

\multirow[c]{2}{35mm}{C-3 (Willingness to use the avatar)}
    & Operator     & 90.4 & 3.8  \\
    & Interlocutor & 83.7 & 2.0  \\
\midrule

O-1 (Achievement of the intended operation)
    & Operator & 73.1 & 13.5 \\
\hdashline[2pt/1pt]

O-2 (Ease of operation)
    & Operator & 92.3 & 0.0  \\
\midrule

I-1 (Human-likeness)
    & Interlocutor & 91.8 & 4.1  \\
\hdashline[2pt/1pt]

I-2 (Ease of conveying emotions and intentions)
    & Interlocutor & 89.8 & 6.1  \\
\bottomrule
\end{tabular}

\vspace{1mm}
\footnotesize{
$^{\dagger}$Positive responses are defined as ratings above the neutral point, and negative responses as ratings below it.
}
\end{table}

The proportions of positive and negative responses relative to the neutral point for each Likert-scale questionnaire item are summarized in Table~\ref{tab:exp1_results_summary}.

Overall, the responses to all questionnaire items were generally positive.  
For C-1, which assessed the comparison with video calls, operators showed a slightly higher proportion of negative responses than interlocutors.  
By contrast, for C-3, which assessed the willingness to use the avatar, approximately 90\% of both operators and interlocutors provided positive responses.

For operator-specific items, the responses were generally positive. However, strong positive responses were limited to O-1, which assessed the degree to which the intended operations were achieved.  
By contrast, for O-2, which assessed the ease of operation, many strong positive responses were observed, and no negative responses were reported.  
For O-3 (see Fig.~\ref{fig:question_o3}), which assessed the physical workload during the operation, the responses ranged from 0 to 78, with a median of 17.5, indicating that many participants reported a relatively low physical workload.

For interlocutor-specific items (I-1 and I-2), consistently high positive response rates were observed.

The Spearman's rank correlation coefficient was used to examine the relationship between the participants' frequency of video calls and their responses to C-1.  
No significant correlations were found for either operators or interlocutors (operators: $r_S = 0.096$, $p = 0.501$, two-sided; interlocutors: $r_S = 0.225$, $p = 0.121$, two-sided).

Furthermore, responses were compared based on whether the participants had prior familiarity with their interlocutors (first-time or previously known).
The results for Questions C-2 and I-2, which include the phrase ``the other person,'' are shown in Fig. ~\ref{fig:known_c2} and Fig. ~\ref{fig:known_i2}, respectively.
For both questions, the responses in both groups were generally skewed toward positive evaluations.
However, for Question C-2, the proportion of ``Strongly felt so'' responses was 8.6\% in the group interacting with a previously known interlocutor, compared with 21.4\% in the first-time interaction group. 
A similar trend was observed for Question I-2. 
The proportion of ``Strongly felt so'' responses was 8.6\% in the previously known group and 35.7\% in the first-time interaction group, indicating higher evaluations among participants interacting with unfamiliar partners.

\paragraph{Implications}
These results suggest that the participants tended to report favorable interaction experiences when using the android avatar.

Across the common items (C-1--C-3), positive evaluations were generally observed for both operators and interlocutors; however, differences were found depending on the aspect being evaluated.
In particular, for the comparison with video calls (C-1), operators showed a relatively higher proportion of negative responses than interlocutors, whereas for the sense of co-presence (C-2), both groups reported consistently positive evaluations. This indicates that participants perceived a sense of sharing the same space with a remote partner, regardless of their role.
In addition, the willingness to use the avatar (C-3) was rated highly by both groups, indicating that the participants maintained a high level of acceptance of the system even when some limitations were perceived in comparison with video calls.
Furthermore, the high evaluations of human-likeness (I-1) and ease of conveying emotions and intentions (I-2) among interlocutors suggest that the avatar may be perceived to be more than a simple interface and as a social entity.
These results indicate that the system can support communication involving non-verbal cues, potentially enhancing perceived interaction quality.

However, regarding the degree to which the intended operations were achieved (O-1), although positive evaluations were observed, strong positive responses were limited, indicating that further improvements in operability are required.
This issue may be influenced not only by limitations in the synchronization system and hardware, but also by differences in latency characteristics across channels, as shown in Appendix~\ref{sec:appendix_latency}.
However, the ease of understanding the operation method (O-2) was rated highly, and the physical workload (O-3) remained relatively low, with a median of 17.5 on a 0--100 scale.
These results suggest that the system is accessible to general users and provides an operational scheme that does not impose a substantial burden.

Finally, in the analysis based on prior familiarity with the interlocutor, positive evaluations of co-presence and emotional communication were observed in both groups, although the proportion of participants with the highest ratings was greater in the first-time interaction group.
These findings indicate that avatar-mediated remote communication can facilitate a strong sense of co-presence and effective emotional exchange regardless of prior relationships between users. 
The findings also hint the system's potential to support the formation of interpersonal relationships between previously unacquainted individuals.

\subsubsection{Experiment 2: Interlocutor impression evaluation}
The results and implications of Experiment 2 are the following.

\begin{figure}[tp]
 \centering
    \begin{tabular}{cc}
        \begin{minipage}{0.5\textwidth}
            \centering                    
            \begin{subfigure}[t]{\columnwidth}
                \centering
                \includegraphics[height=4.5cm]{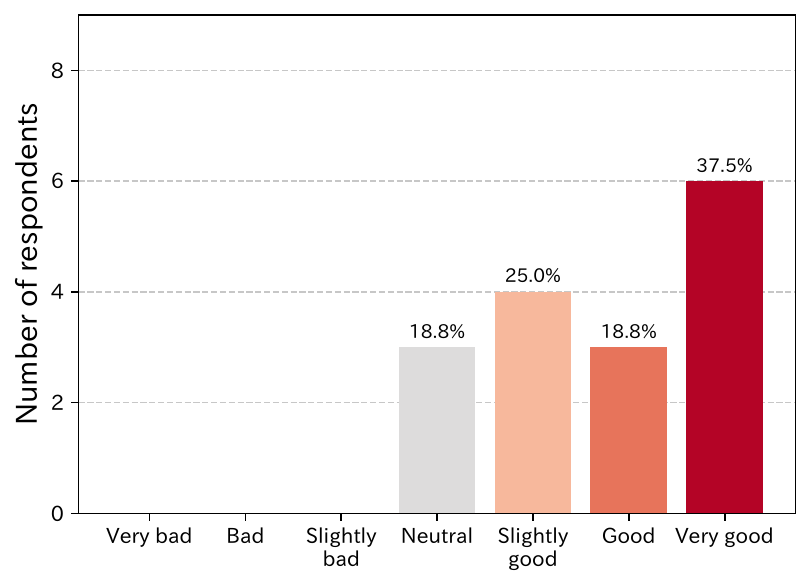}
                \subcaption{Question C-1: How did you find this remote interaction experience compared to a video call?}
                \label{fig:exp2_c1}
            \end{subfigure}
        \end{minipage}
        &
        \begin{minipage}{0.5\textwidth}
            \centering 
            \begin{subfigure}[t]{\columnwidth}
                \centering
                \includegraphics[height=4.5cm]{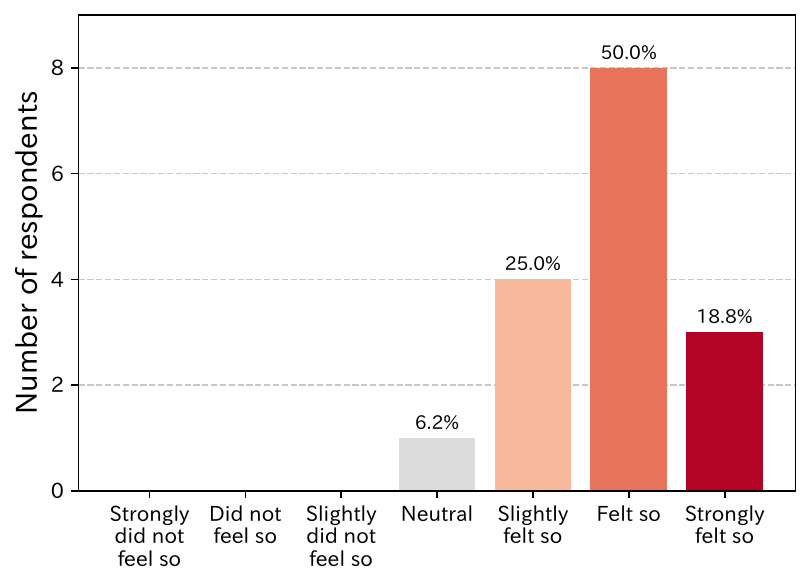}
                \subcaption{Question C-2: Did you feel as if the other person was in the same physical space as you?\\}
                \label{fig:exp2_c2}
            \end{subfigure}
        \end{minipage}
    \end{tabular}
    \\
    \begin{tabular}{cc}
        \begin{minipage}{0.5\textwidth}
            \centering                    
            \begin{subfigure}[t]{\columnwidth}
                \centering
                \includegraphics[height=4.5cm]{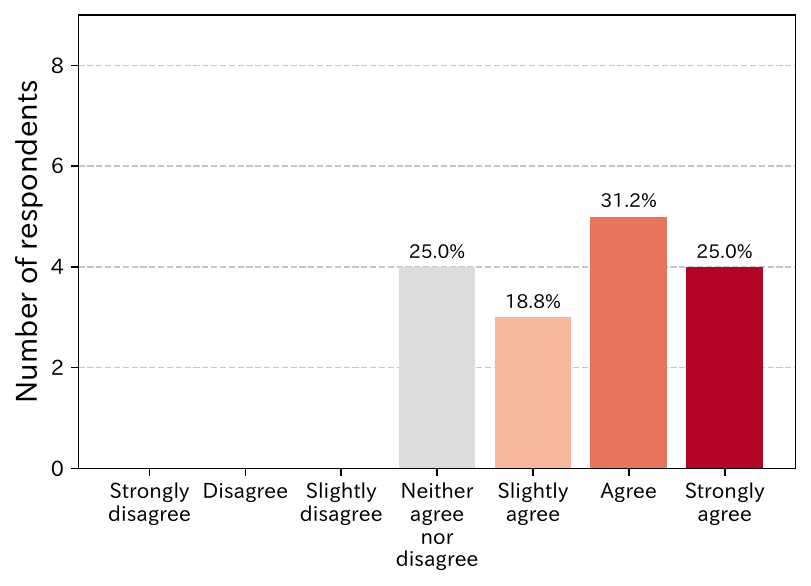}
                \subcaption{Question C-3: Would you like to use this type of robot-avatar-based remote communication tool?}
                \label{fig:exp2_c3}
            \end{subfigure}
        \end{minipage}
        &
        \begin{minipage}{0.5\textwidth}
            \centering 
            \begin{subfigure}[t]{\columnwidth}
                \centering
                \includegraphics[height=4.5cm]{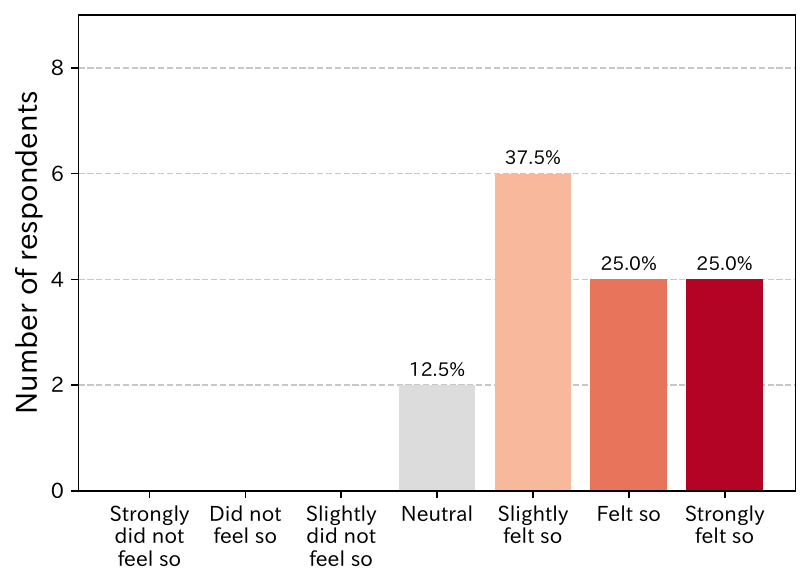}
                \subcaption{Question I-1: Did the robot avatar's responses and behaviors feel human-like?\\}
                \label{fig:exp2_i1}
            \end{subfigure}
        \end{minipage}
    \end{tabular}
    \\
    \begin{subfigure}[t]{\columnwidth}
        \centering
        \includegraphics[height=4.5cm]{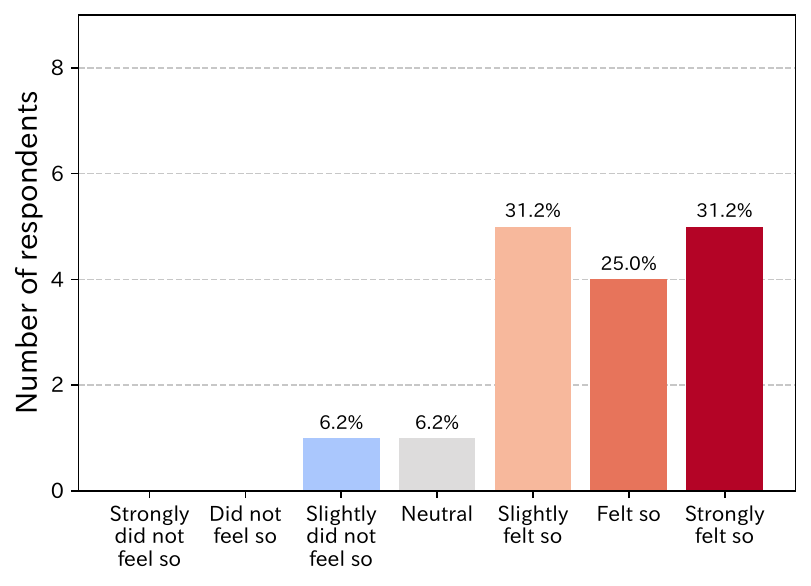}
        \subcaption{Question I-2: Did the other person's emotions and intentions feel well conveyed through the robot avatar?}
        \label{fig:exp2_i2}
    \end{subfigure}
    \caption{Distribution of questionnaire responses in Experiment 2 ($N=16$）.}
    \label{fig:exp2_questions}
\end{figure}

\paragraph{Results}
The results for questions C-1, C-2, C-3, I-1, and I-2 are shown in Fig. ~\ref{fig:exp2_c1}, Fig.~\ref{fig:exp2_c2}, Fig.~\ref{fig:exp2_c3}, Fig.~\ref{fig:exp2_i1} and Fig.~\ref{fig:exp2_i2}, respectively.

\begin{table}[t]
\renewcommand{\arraystretch}{1.2}
\centering
\caption{Proportions of positive and negative responses for each questionnaire item in Experiment 2}
\label{tab:exp2_results_summary}
\begin{tabular}{>{\raggedright\arraybackslash}m{35mm} >{\raggedleft\arraybackslash}m{20mm} >{\raggedleft\arraybackslash}m{20mm}}
\toprule
Question & \multicolumn{1}{m{20mm}}{Positive response$^{\dagger}$ (\%)} & \multicolumn{1}{m{20mm}}{Negative response$^{\dagger}$ (\%)} \\
\midrule
C-1 (Comparison with video calls) & 81.3 & 0.0 \\
\hdashline[2pt/1pt]
C-2 (Sense of co-presence) & 93.8 & 0.0 \\
\hdashline[2pt/1pt]
C-3 (Willingness to use the avatar) & 75.0 & 0.0 \\
\midrule
I-1 (Human-likeness) & 87.5 & 0.0 \\
\hdashline[2pt/1pt]
I-2 (Ease of conveying emotions and intentions) & 87.5 & 6.2 \\
\bottomrule
\end{tabular}
\vspace{1mm}
\footnotesize{
$^{\dagger}$Positive responses are defined as ratings above the neutral point, and negative responses as ratings below it.
}
\end{table}

The proportions of positive and negative responses relative to the neutral point for each questionnaire item are summarized in Table~\ref{tab:exp2_results_summary}.  
Overall, the responses tended to be positive across all items, with very few negative responses.  
In particular, for C-1, which assessed the quality of the experience compared to video calls, 37.5\% of participants selected ``Very good,'' representing the largest proportion of responses.

The Spearman's rank correlation coefficient was used to examine the relationship between the participants' frequency of video calls and their responses to C-1.  
No significant correlation was observed ($r_S = -0.081$, $p = 0.767$, two-sided).

\paragraph{Implications}
These results suggest that when the avatar was operated by a trained operator, interlocutors tended to report more favorable interaction experiences than video calls.
No negative responses were observed across any evaluation item, and responses were strongly skewed toward positive ratings. This pattern suggests that, when properly operated, the system can deliver a consistently positive interaction experience.

In particular, the high evaluations of human-likeness (I-1) and ease of conveying emotions and intentions (I-2) suggest that the participants perceived non-verbal information to be effectively conveyed through the system.
Furthermore, the high ratings for the sense of co-presence (C-2) prove that the participants felt a sense of sharing the same space with a remote operator.

Additionally, the high ratings for experience quality relative to video calls (C-1) and  willingness to use the avatar (C-3) signify that the system has the potential to serve as a viable alternative to existing remote communication methods.

Nevertheless, because these results were obtained in the presence of a trained operator, caution should be exercised when generalizing these findings to general users.
However, given that Experiment 1 also showed generally positive evaluations, the system appears capable of delivering a satisfactory interaction experience to a broad range of users, although outcomes still depend in part on operator proficiency.

Thus, Study 3 achieved its objective of exploring the experiences and impressions of general participants interacting through the android avatar in real-world settings.

\section{Discussion}
This section summarizes the findings obtained from Studies 1 to 3.

\subsection{Summary of Findings}
This study investigated the feasibility and effectiveness of a remote communication system using an android avatar through a multifaceted evaluation consisting of a long-term deployment at an Expo (Study 1), an experiential event at an elementary school (Study 2), and evaluation experiments with the general public (Study 3).

Study 1 demonstrated that the system could be operated stably over an extended period in a public environment.  
Although several technical issues occurred during deployment, continuous exhibition was achieved through maintenance and adjustments to the operation strategy, indicating the practical feasibility of android avatars in in-the-wild settings.

Study 2 confirmed that remote communication using the proposed system can be intuitively understood and successfully established, even by users without specialized knowledge, through experiential activities involving elementary school students.
In particular, the observation that children asked questions and interacted with a remotely located operator via the avatar demonstrates that the avatar can be naturally accepted as an interaction partner and recognized as a social entity, even by young users.

Study 3 showed that the system received consistently high subjective evaluations from general participants across multiple aspects, including the experience quality relative to video calls, sense of co-presence, willingness to use the avatar, ease of operation, human likeness, and the transmission of emotions and intentions.
These findings indicate that the system is acceptable to general users and has the potential to be used as a remote communication tool in real-world settings.
In addition, when the avatar was operated by a trained operator, the evaluations tended to be higher, suggesting that the system may provide a positive interaction experience when appropriately operated.

Collectively, these results demonstrate that the proposed system achieves consistently positive outcomes across long-term deployment in public spaces, experiential use by general users, and quantitative evaluations.
This is comprehensive evidence of the feasibility and effectiveness of remote communication mediated by android avatars.

These findings are consistent with prior in-the-wild studies suggesting that android robots can be accepted in public or semi-public service environments.
Umetani et al.~\cite{TUmetani2019} reported the feasibility of a remote reference-desk service using an android robot in a university library through long-term operation.
Heisler and Becker-Asano~\cite{MHeisler2025} similarly found that visitors to a public museum generally perceived a conversational android positively and often interacted with it out of curiosity.
The present study complements and extends these findings by showing that a full-body android avatar can be deployed over an extended period and can support positive interaction experiences across multiple real-world contexts, including a long-term public exhibition, educational exchange, and public interaction studies.

\subsection{Implications for Social Robotics}
From the perspective of social robotics, these results suggest that the physical embodiment of android avatars may play an important role in remote communication.  
This is consistent with prior studies demonstrating that physical embodiment enhances social telepresence~\cite{KTanaka2014} and that physically present robots elicit higher levels of trust and social engagement than on-screen agents~\cite{WABainbridge2008}.  

In Study 3, many participants reported positive evaluations of the quality of the interaction experience compared to video calls, and high ratings were observed for both the sense of co-presence and ease of conveying emotions and intentions.  
These findings hint that the system can contribute to a sense of spatial and social presence in remote communication.  
Furthermore, as videoconferencing has been reported to provide insufficient social telepresence than face-to-face interaction~\cite{KTanaka2014}, the present results suggest that incorporating physical embodiment may help alleviate such limitations.
The results of Studies 1 and 2 also show that android avatars can be accepted and operated not only in controlled laboratory settings but also in real-world environments, including public spaces and educational contexts.
The successful establishment of natural interactions in settings involving diverse users, including general visitors and children, provides important insights into the social deployment of android avatars.

Additionally, this study is significant because it comprehensively examines the system's operational feasibility and social acceptance through field-based experiments conducted in real-world settings rather than in controlled laboratory environments.
These findings can be regarded as foundational knowledge that supports future comparative studies and theoretical investigations in this field.

\subsection{Guidelines for Future Avatar Systems}
The findings of this study have several implications for the design of future avatar systems.

First, to achieve long-term deployment in public environments, robust and maintainable system design that can handle unexpected failures is essential.
As observed in Study 1, unforeseen technical issues are unavoidable. Systems should therefore be designed for maintainability that enables rapid intervention, along with mechanisms for automatic recovery from temporary disruptions such as communication failures.
From a hardware-design perspective, many android platforms have relied on pneumatic actuation because of its quiet and compliant motion characteristics~\cite{HIshiguro2005,MHeisler2025,Nishio2007,Thalmann2017}.
However, such pneumatic systems also require air-supply infrastructure and careful installation in public environments.
In contrast, Yui was implemented as an electrically actuated full-body system with subsystem-level control interfaces.
Although failures still occurred during the Expo deployment, the results suggest that subsystem-based design and rapid component-level recovery are important for long-term public operation.

Second, a hybrid operation that combines autonomous functions with teleoperation is proposed to be effective for sustainable deployment in real-world settings.
Operations that rely solely on human operators are impractical from the perspectives of cost and workload.
Therefore, adopting an operational strategy that allows flexible switching between autonomous behavior and teleoperation depending on the situation is crucial.

Third, the importance of interface design tailored to user characteristics and usage contexts was demonstrated.
The results of Studies 2 and 3 indicate that intuitive interfaces that can be used by general users and children are necessary, whereas operations by trained users enable higher-quality interaction experiences.
Accordingly, future systems should allow for the selection and adaptation of appropriate interfaces based on user proficiency and intended use.

\subsection{Limitations}
This study had several limitations.

First, as this study focused on system integration and the verification of feasibility in real-world settings, no strictly controlled comparative experiments with other robot avatars or alternative interfaces were conducted.
Second, Studies 1 and 2 included qualitative observations, and their interpretation may involve some degree of subjectivity. However, these studies were conducted in real-world environments, such as an Expo and educational settings, where collecting controlled quantitative data was inherently difficult due to practical constraints and ethical considerations related to participant experience.  
Therefore, the qualitative findings should be interpreted as valuable observations in a real-world context.

Third, the participants in Study 3 were recruited from event visitors, which may have introduced a bias toward individuals with a relatively high interest in technology.  
A possibility is that this bias contributed to the generally positive evaluations.  
However, unlike laboratory-based experiments, this study was conducted on a diverse group of general visitors with a wide range of ages and backgrounds.  
The consistently high evaluations observed in this context suggest that the proposed system has the potential to be accepted in real-world settings.

As a supplementary point, as shown in Appendix~\ref{sec:appendix_latency}, communication delays and differences in latency characteristics across channels may also have influenced the operability and interaction experience.
In this study, such latency-related issues were tolerated because the primary objective was to demonstrate the concept presented in Section~\ref{sec21} and to investigate the feasibility of operating the system in real-world environments; however, they should be regarded as one possible supplementary factor when interpreting the evaluation results.

\subsection{Future direction}
In future studies, a high-priority issue will be to more clearly position the characteristics of the proposed system through comparisons with other remote communication methods and different types of avatar systems.

In addition, when conducting research on the social deployment of android avatars, an essential point will be to establish new guidelines for privacy protection and informed consent, particularly in real-world settings, such as public spaces and educational environments that involve unspecified participants and minors. These guidelines should address both technical and operational aspects.
Furthermore, improving the operation interface and reducing latency so that general users can achieve interaction quality comparable to that of trained users will be key challenges for practical deployment.

\section{Conclusion}
This study explored the practical feasibility, effectiveness, and social acceptance of an android avatar as a medium for remote communication through the system integration of the android avatar ``Yui,'' long-term deployment in a real-world environment, and exploratory evaluations with the general public.

First, through an approximately six-month exhibition at Expo 2025 in Osaka, Kansai, we demonstrated that the system operated over an extended period in a public environment.
Although several technical issues occurred during operation, continuous deployment was maintained through maintenance and adjustments, indicating the system's practicality and a degree of robustness in real-world settings.
Through an experiential event with elementary school students and evaluation experiments with the general public, the system was found to be intuitive and capable of supporting interaction even for users without specialized knowledge. 
In particular, consistently positive evaluations given by users of diverse ages and backgrounds suggest that the system has the potential for acceptance in real-world contexts.

Subjective evaluations further indicated that participants reported more favorable interaction experiences than that with video calls in terms of satisfaction, sense of co-presence, human likeness, and emotional communication. 
These findings suggest that physical embodiment, as realized by android avatars, may contribute to perceived improvements in the quality of remote communication.
Overall, this study demonstrated the feasibility and effectiveness of remote communication mediated by android avatars in real-world environments and provided foundational insights for their social deployment.
Future work should further examine the system across a wider range of usage scenarios, clarify its effectiveness through comparisons with other types of robot avatars and communication interfaces, and establish design guidelines suitable for practical deployment.

\backmatter

\bmhead{Acknowledgements}
This work was supported by JST Moonshot R\&D, Grant Number JPMJMS2011 (overall research, system integration, field deployment, and evaluation), and in part by a joint research agreement originally established with A-Lab, Inc. (Japan), now part of AVITA, Inc. (Japan), under Project Code YC2022077 (full-body expression of the android avatar).

The authors thank all the participants and cooperating institutions for supporting the field deployments and evaluations. We are grateful to the staff and collaborators of Expo 2025 Osaka, Kansai (Signature Pavilion ``Future of Life'') for operational and logistical support during the long-term in-the-wild deployment (April--October 2025). We also thank the members of the NAKATA Lab, the University of Electro-Communications, for their assistance with system preparation and on-site operations during field deployment. We also thank the cooperating schools and teachers for supporting the educational exchange study, as well as the organizers and venue staff in Ishikawa for facilitating public evaluations.

We thank Professor Hiroshi Ishiguro (Theme Producer, Expo 2025 Osaka, Kansai, Japan; Signature Pavilion ``Future of Life'') for his support. We thank Mr. Tatsuya Matsui and colleagues at Flower Robotics Inc. for the exterior design of the Expo version of Yui, Mr. Hitoshi Maida for designing the Kaga-Yuzen kimono, and secca inc. for coordinating the Wajima-nuri finish of Yui's hair piece. We also thank Mr. Kohei Takatsuki and colleagues at A-Lab Co., Ltd. for their engineering support in the development of Yui's Head Unit.

We thank Mr. Takashi Tokuda and colleagues at Keigan Inc. for engineering support in the development of the custom motors; Sumitomo Heavy Industries, Ltd. for technical support related to cycloidal drives; and Mr. Takeshi Maeda and colleagues at Vstone Co., Ltd. for engineering support.

\section*{Declarations}

\begin{itemize}
\item Funding: This work was supported by JST Moonshot R\&D Grant Number JPMJMS2011.
\item Competing interests: The authors have no competing interests to declare that are relevant to the content of this article.
\item Ethics approval and consent to participate: The study protocol was approved by the Ethics Committee of the University of Electro-Communications (No. H25054). Informed consent was obtained from all participants prior to participation.
\item Consent for publication: Informed consent for publication of identifiable images was obtained from the individuals shown in the manuscript.
\item Data availability: The data supporting the findings of this study are available from the corresponding author upon reasonable request.
\item Materials availability: The android avatar system (Yui) and related hardware components used in this study are research prototypes and are not publicly available.
\item Code availability: The code used in this study is not publicly available.
\item Author contribution: Author contributions: KS: Conceptualization, Methodology, Software, Validation, Formal analysis, Investigation, Resources, Data curation, Writing -- original draft, Writing -- review \& editing, Visualization. MN: Conceptualization, Methodology, Software, Validation, Resources, Writing -- review \& editing, Supervision. TM: Methodology, Software, Validation, Resources, Writing -- original draft, Visualization. YN: Conceptualization, Methodology, Validation, Investigation, Resources, Writing -- original draft, Writing -- review \& editing, Visualization, Supervision, Project administration, Funding acquisition.
\end{itemize}

\begin{appendices}

\section{Study 1: Avatar Appearance}
\label{sec:appendix_avatar}

\begin{figure}[t]
  \centering
  \includegraphics[width=0.8\linewidth]{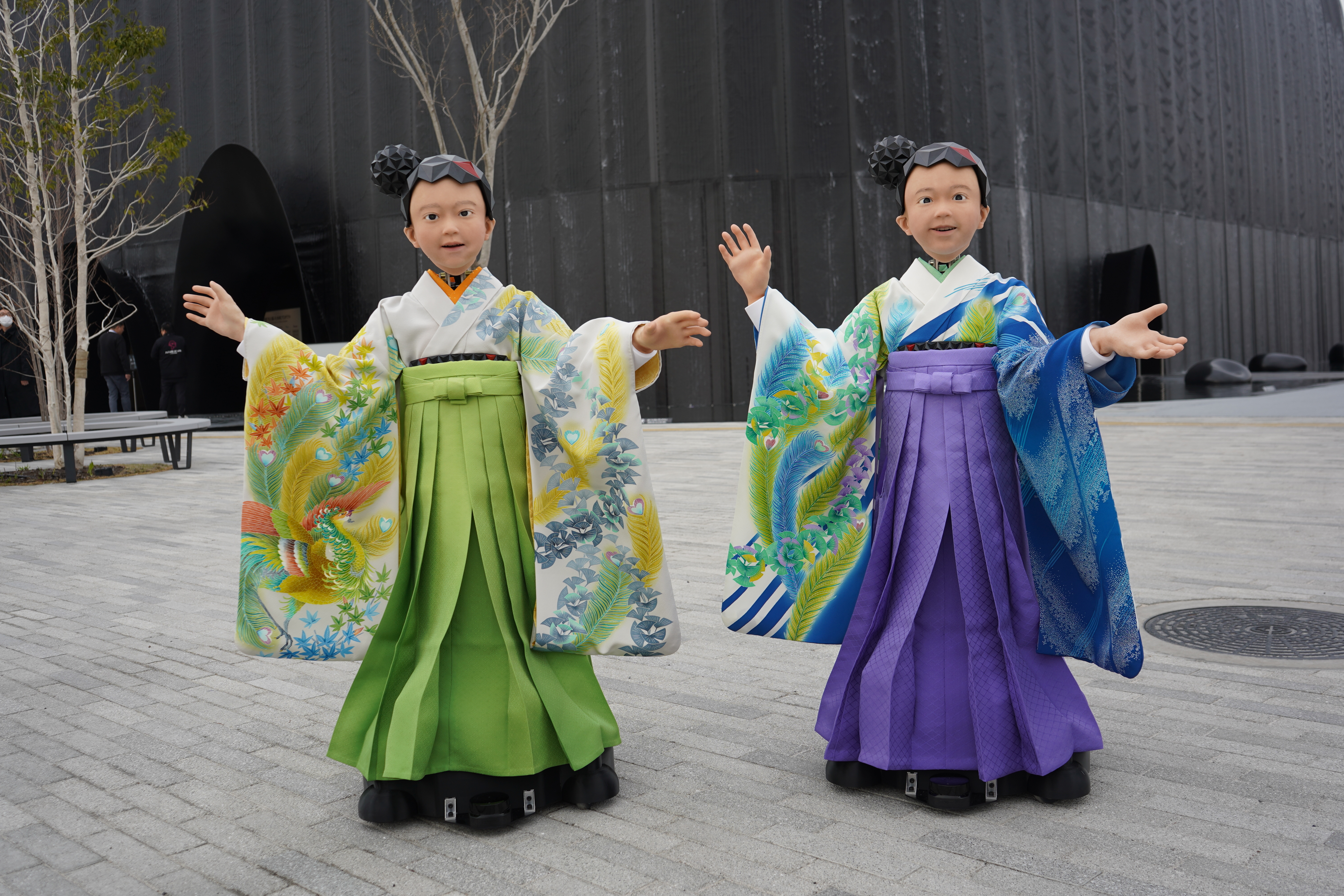}
  \caption{Two android avatar units used in the long-term deployment.}
  \label{fig:yui_01_02}
\end{figure}

Two android avatars were prepared for long-term deployment.  
Fig.~\ref{fig:yui_01_02} shows both units used in the exhibition.

Both avatars wore traditional Japanese attire consisting of a kimono and hakama, with head decorations inspired by traditional Japanese hairstyles and waist decorations resembling the kimono obi.  
Costumes were designed in a gender-neutral style, combining elements typically associated with feminine kimonos and masculine hakama.

In addition, the design incorporated traditional crafts associated with Ishikawa Prefecture in response to the 2024 Noto Peninsula Earthquake.  
Specifically, the kimono featured Kaga Yuzen, a traditional dyeing technique from Ishikawa, while the head and obi decorations were finished with a lacquer coating using Wajima-nuri, a traditional lacquerware technique from Ishikawa.

\section{Study 3: Participants Distribution}\label{sec:appendix_dist}

\begin{figure}[t]
 \centering
    \begin{tabular}{cc}
        \begin{minipage}{0.5\textwidth}
            \centering                    
            \begin{subfigure}[t]{\columnwidth}
                \centering
                \includegraphics[height=4.8cm]{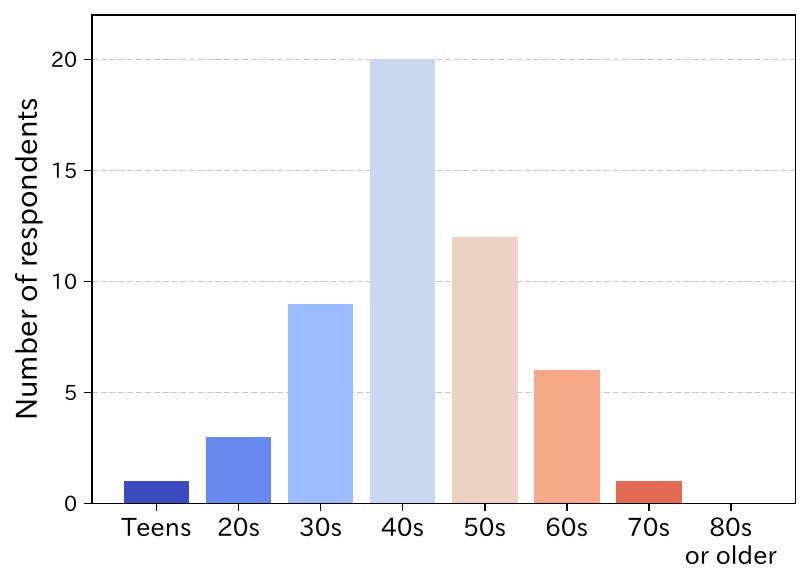}
                \subcaption{Operators ($N=52$)}
                \label{fig:operator_participants}
            \end{subfigure}
        \end{minipage}
        &
        \begin{minipage}{0.5\textwidth}
            \centering 
            \begin{subfigure}[t]{\columnwidth}
                \centering
                \includegraphics[height=4.8cm]{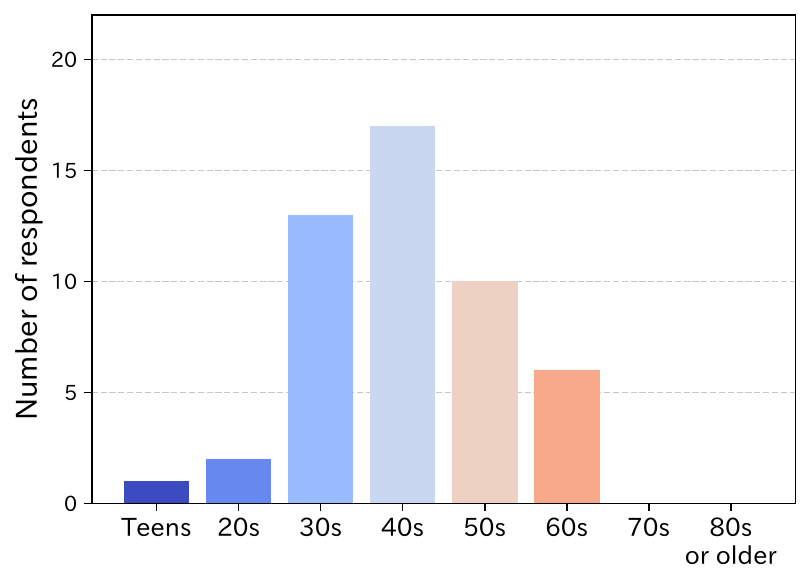}
                \subcaption{Interlocutors ($N=49$)}
                \label{fig:interlocutor_participants}
            \end{subfigure}
        \end{minipage}
    \end{tabular}
    \caption{Age distributions of operators and interlocutors in Experiment 1.}
    \label{fig:dist_exp1}
\end{figure}

\begin{figure}[t]
  \centering
  \includegraphics[width=0.8\textwidth]{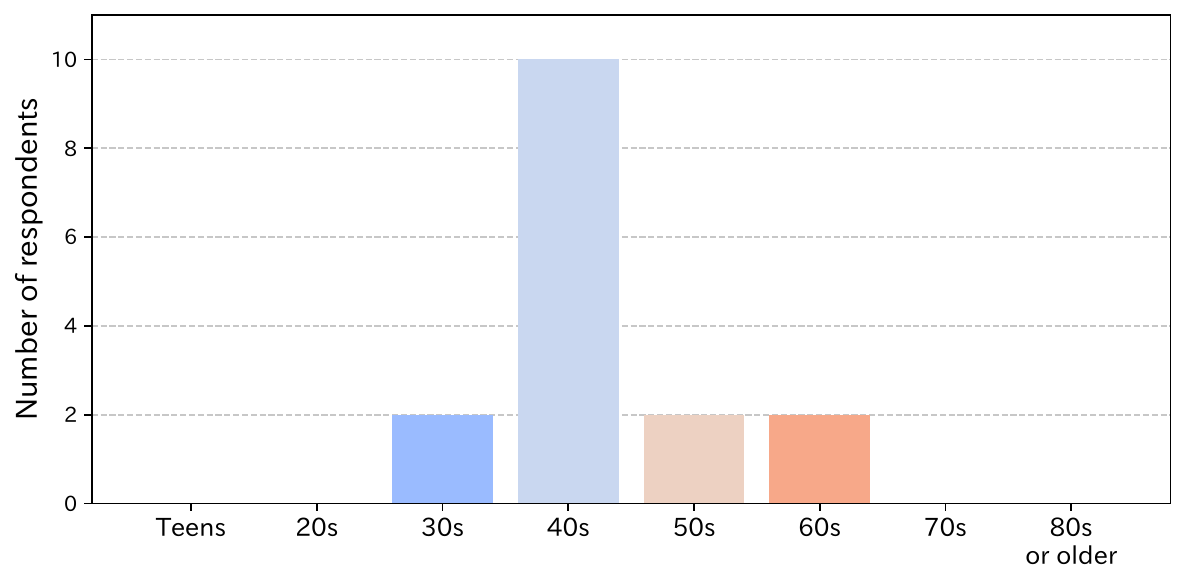}
  \caption{Age distribution in Experiment 2 ($N=16$)}
  \label{fig:dist_exp2}
\end{figure}

The age distributions of participants are shown in Fig.~\ref{fig:dist_exp1} and Fig.~\ref{fig:dist_exp2}.  
Participants reported their age in decade-based categories (e.g., teens, 20s, 30s).

\section{Latency Measurement}\label{sec:appendix_latency}

This appendix reports the results of end-to-end latency measurements for the video, audio, and motion channels during teleoperation under laboratory conditions.
These measurements were not intended to strictly reproduce the exact latency conditions of the experiments reported in this study, but rather to characterize the latency properties of the proposed system.

During the measurements, the operator-side PC and the avatar-side PC were connected to different mobile networks to avoid communication within the same local area network.
In addition, to monitor network conditions during the measurements, round-trip time (RTT) to an external server and packet loss were recorded using the ping command for each latency measurement of the individual channels.
The results are summarized in Table~\ref{tab:rtt_summary}. No packet loss was observed in any of the measurements.

\begin{table*}[t]
\renewcommand{\arraystretch}{1.2}
\centering
\caption{Round-trip time (RTT) statistics during latency measurements.}
\label{tab:rtt_summary}
\begin{tabular}{c|ccc|ccc}
\hline
& \multicolumn{3}{c|}{Operator side} & \multicolumn{3}{c}{Avatar side} \\
Channel
& Min [s]
& Max [s]
& Mean [s]
& Min [s]
& Max [s]
& Mean [s]
\\
\hline

Video
& 0.028 & 0.072  & 0.041
& 0.026 & 0.095  & 0.046
\\

Audio
& 0.028 & 0.062 & 0.039
& 0.020 & 0.153 & 0.041
\\

Motion
& 0.026 & 0.223 & 0.041
& 0.019 & 0.222 & 0.036
\\

\hline
\end{tabular}
\end{table*}

\subsection{Video Feedback Latency}

The video latency was measured as the delay between the image captured by the avatar's eye camera and its display on the operator-side screen.
A digital timer displaying milliseconds was shown on a display and positioned within the avatar's field of view.
At the same time, the video received from the avatar's eye camera was displayed on the operator-side PC screen.
The physical timer display and the timer displayed within the received video on the operator-side PC screen were simultaneously recorded using a single camera, and the difference between the displayed times was calculated as the video latency.
The recording was performed at \qty{120}{fps} for approximately one minute.

Ten frames were extracted at 5-s intervals starting 5 \qty{120}{s} after the beginning of the recording, and the time difference between the physical timer and the timer shown in the received video was measured for each frame.

The results are summarized in Table~\ref{tab:visual_latency}. 
The mean video feedback latency was \qty{0.45}{s}.

\begin{table}[t]
\centering
\renewcommand{\arraystretch}{1.2}
\caption{Measured video feedback latency.}
\label{tab:visual_latency}
\begin{tabular}{c@{\hspace{1.2cm}}c}
\hline
Frame & Delay [s] \\
\hline
1  & 0.455 \\
2  & 0.455 \\
3  & 0.449 \\
4  & 0.447 \\
5  & 0.458 \\
6  & 0.453 \\
7  & 0.425 \\
8  & 0.437 \\
9  & 0.427 \\
10 & 0.459 \\
\hline
Mean $\pm$ SD & $0.447 \pm 0.013$\\
\hline
\end{tabular}
\end{table}

\subsection{Audio Transmission Latency}

Since audio was transmitted bidirectionally, audio latency was measured separately for transmission from the operator side to the avatar side and from the avatar side to the operator side.

The latency from the operator side to the avatar side was measured as the time difference between the playback of a sound on the operator side and its reproduction through the speaker on the avatar side.
The avatar was placed in the same room as the operator side, and a short reference audio signal was played near the operator-side microphone while the reproduced sound from the avatar-side speaker was recorded.
As the reference signal, a \qty{500}{Hz} sinusoidal tone with a duration of \qty{0.1}{s} was used and played 10 times consecutively at \qty{5}{s} intervals.

Subsequently, to measure the latency from the avatar side to the operator side, the same reference signal was played near the avatar-side microphone, and the sound reproduced from the operator-side speaker was recorded.

For the recorded audio data, ten 5-s segments segments were extracted so that each segment contained both the original reference signal and the corresponding reproduced sound.
For each segment, delay estimation based on autocorrelation was performed, and the time shift that produced the maximum correlation within the range of 0.1--\qty{2.0}{s} was identified.
This range was chosen to exclude shifts shorter than the duration of the reference signal itself while remaining sufficiently longer than the expected transmission delay.
The mean value across the 10 segments was used as the audio latency.

The results are summarized in Table~\ref{tab:audio_latency}.
The mean audio transmission latency was \qty{1.12}{s} from the operator side to the avatar side and \qty{0.34}{s} from the avatar side to the operator side.

\begin{table}[t]
\centering
\renewcommand{\arraystretch}{1.2}
\caption{Measured audio transmission latency.}
\label{tab:audio_latency}
\begin{tabular}{ccc}
\hline
Segment & Operator $\rightarrow$ Avatar [s] & Avatar $\rightarrow$ Operator [s] \\
\hline
1  & 1.119 & 0.312 \\
2  & 1.118 & 0.347 \\
3  & 1.120 & 0.330 \\
4  & 1.118 & 0.332 \\
5  & 1.119 & 0.330 \\
6  & 1.120 & 0.348 \\
7  & 1.124 & 0.346 \\
8  & 1.120 & 0.348 \\
9  & 1.120 & 0.348 \\
10 & 1.120 & 0.348 \\
\hline
Mean $\pm$ SD & 1.120 $\pm$ 0.002 & 0.339 $\pm$ 0.012 \\
\hline
\end{tabular}
\end{table}

\subsection{Motion Synchronization Latency}

Motion latency was measured as the delay from the operator's body movement to the corresponding physical movement of the avatar.
In this study, rather than measuring the delay of each individual joint, we measured the end-to-end latency using representative movements that were relatively easy to evaluate for the head and arm, which are controlled by different control systems.
Specifically, jaw opening/closing and arm lifting/lowering were used as representative movements.

For the jaw opening/closing movement, motion-capture markers were attached near the lower lip of both the operator and the avatar, and the operator periodically opened and closed their jaw under normal teleoperation conditions.
For the arm lifting/lowering movement, motion-capture markers were attached to the left wrist of both the operator and the avatar, and the operator periodically lifted and lowered their left arm under normal teleoperation conditions.
Each movement was recorded for one minute, during which the operator repeatedly performed the target movement.

For the analysis, the vertical coordinates of the operator-side and avatar-side markers were used.
A 50-s segment from 5 s after the start of recording to 5 s before the end of recording was extracted.
The vertical trajectories of the operator-side and avatar-side markers were normalized, and cross-correlation was then computed.
The time shift that produced the maximum correlation within a delay range of up to \qty{2}{s} was estimated as the motion synchronization latency.
This method evaluated the end-to-end motion latency, including tracking, communication, control processing, and actuator response, rather than the response delay of individual motors.

As a result, the motion synchronization latency was \qty{0.38}{s} for jaw opening/closing and \qty{0.87}{s} for left arm lifting/lowering.

\subsection{Summary}

Table~\ref{tab:latency_summary} summarizes the measured latency for each channel. The audio transmission latency from the operator to the avatar was larger than in the reverse direction, presumably because the operator-to-avatar path involved real-time voice conversion.

Regarding motion, the latency discrepancy between the head (jaw movement) and arm (lifting/lowering) may be attributable to differences in communication protocols, buffering times, maximum speed settings, and whether the command transmission process from the avatar-side Unity application to the control program was required.

In this study, the primary objective was to demonstrate the concept and examine the feasibility of system operation in real-world environments. Consequently, the experiments were conducted while acknowledging certain degrees of latency and inter-channel variability. However, to facilitate smoother remote interaction, future work should focus on reducing latency and improving synchronization across all modalities, including communication, audio-visual processing, and the control systems of individual robot components.

\begin{table}[t]
\centering
\renewcommand{\arraystretch}{1.2}
\caption{Summary of end-to-end latency measurements.}
\label{tab:latency_summary}
\begin{tabular}{lc}
\hline
Channel & Latency [s] \\
\hline
Video feedback & 0.45 \\
Audio: Operator $\rightarrow$ Avatar & 1.12\\
Audio: Avatar $\rightarrow$ Operator & 0.34 \\
Motion: Head (jaw opening) & 0.38 \\
Motion: Arm (left arm lifting) & 0.87 \\
\hline
\end{tabular}
\end{table}

\end{appendices}

\bibliography{sn-bibliography}

\end{document}